%% file: clear2024.tex
\documentclass[final,12pt]{clear2024} 

\title[Toward the Identifiability of Comparative Deep Generative Models]{Toward the Identifiability of Comparative Deep Generative Models}
\usepackage{times}

\usepackage[utf8]{inputenc} 
\usepackage[T1]{fontenc} 
\usepackage{url}  
\usepackage{booktabs} 
\usepackage{amsfonts} 
\usepackage{nicefrac} 
\usepackage{microtype}  
\usepackage{xcolor} 
\usepackage{tikz}
\usepackage{multirow}
\usepackage{thmtools}

\usetikzlibrary{bayesnet}
\usetikzlibrary{arrows}
\usepackage{color}
\usepackage{graphicx}
\usetikzlibrary{backgrounds}
\usepackage{float}
\usepackage{bm}
\usepackage{wrapfig}

\usepackage{mathtools}
\DeclarePairedDelimiterX{\infdivx}[2]{(}{)}{%
  #1\;\delimsize\|\;#2%
}

\newcommand{\Prob}{\mathbb{P}}
\newcommand{\E}{\mathbb{E}}
\newcommand{\R}{\mathbb{R}}
\newcommand{\C}{\mathbb{C}}
\newcommand\norm[1]{\lVert#1\rVert}
\newtheorem{theo}{Theorem}
\newtheorem{lem}{Lemma}

\newcommand\restr[2]{{
  \left.\kern-\nulldelimiterspace 
  #1 
  \littletaller 
  \right|_{#2} 
  }}

\newcommand{\littletaller}{\mathchoice{\vphantom{\big|}}{}{}{}}
\newcommand{\x}{\bm{x}}
\newcommand{\z}{\bm{z}}
\newcommand{\s}{\bm{s}}
\newcommand{\y}{\bm{y}}
\newcommand{\tv}{\bm{t}}
\newcommand{\ta}{\bm{t^1}}
\newcommand{\tb}{\bm{t^2}}
\newcommand{\uvar}{\bm{u}}
\newcommand{\vv}{\bm{v}}
\newcommand{\zero}{\bm{0}}
\newcommand{\equald}{\smash{\,{\buildrel d \over =}\,}}
\newcommand{\indep}{\perp \!\!\! \perp}

\usepackage{amssymb}
\usepackage{amsmath,amscd}
\usepackage{color}

\DeclareMathOperator*{\argmin}{arg\,min}

\usepackage{titletoc}
\usepackage{hyperref}
\usepackage[page,header]{appendix}
\usepackage{mathtools, nccmath}

\newcommand*{\QEDB}{\null\nobreak\hfill\ensuremath{\square}}%

\clearauthor{\Name{Romain Lopez} \Email{lopez.romain@gene.com} \\
\addr Genentech Research and Early Development\\
\addr Stanford University \\
\AND
\Name{Jan-Christian Huetter} \Email{huettej1@gene.com} \\
\addr Genentech Research and Early Development\\
\AND
\Name{Ehsan Hajiramezanali} \Email{hajiramm@gene.com}\\
\addr Genentech Research and Early Development\\
\AND
\Name{Jonathan K. Pritchard} \Email{pritch@stanford.edu} \\
\addr Stanford University\\
\AND
\Name{Aviv Regev} \Email{regeva@gene.com}\\ 
\addr Genentech Research and Early Development
}

\begin{document}
\maketitle
\input{0_abstract}


\input{1_introduction}
\input{2_background}

\input{3_identifiability_theory}
\input{4_impact_regularization}

\input{5_algorithm}
\input{6_experiments}
\input{8_discussion}

\acks{We thank S\'ebastien Lachapelle for providing insights and early guidance through the conception of this work. We acknowledge Kelvin Chen, Taka Kudo for discussions about modeling single-cell perturbation data sets. We thank Jeffrey Spence, Hanchen Wang as well as Saeed Saremi for feedback on this manuscript.

Disclosures: Romain Lopez, Jan Christian Huetter, Ehsan Hajiramezanali and Aviv Regev are employees of Genentech, and / or have equity in Roche. Jonathan Pritchard acknowledges support from grant R01HG008140 from the National Human Genome Research Institute. Aviv Regev is a co-founder and equity holder of Celsius Therapeutics and an equity holder in Immunitas. She was an SAB member of ThermoFisher Scientific, Syros Pharmaceuticals, Neogene Therapeutics, and Asimov until July 31st, 2020.}

\bibliography{clear2024}

\newpage
\appendix

\input{supplements}

\end{document}

%% file: 0_abstract.tex
\begin{abstract}
Deep Generative Models (DGMs) are versatile tools for learning data representations while adequately incorporating domain knowledge such as the specification of conditional probability distributions. Recently proposed DGMs tackle the important task of comparing data sets from different sources. One such example is the setting of contrastive analysis that focuses on describing patterns that are enriched in a target data set compared to a background data set. The practical deployment of those models often assumes that DGMs naturally infer interpretable and modular latent representations, which is known to be an issue in practice. 
Consequently, existing methods often rely on ad-hoc regularization schemes, although without any theoretical grounding. Here, we propose a theory of identifiability for comparative DGMs by extending recent advances in the field of non-linear independent component analysis. We show that, while these models lack identifiability across a general class of mixing functions, they surprisingly become identifiable when the mixing function is piece-wise affine (e.g., parameterized by a ReLU neural network). 
We also investigate the impact of model misspecification, and empirically show that previously proposed regularization techniques for fitting comparative DGMs help with identifiability when the number of latent variables is not known in advance. Finally, we introduce a novel methodology for fitting comparative DGMs that improves the treatment of multiple data sources via multi-objective optimization and that helps adjust the hyperparameter for the regularization in an interpretable manner, using constrained optimization. We empirically validate our theory and new methodology using simulated data as well as a recent data set of genetic perturbations in cells profiled via single-cell RNA sequencing.
\end{abstract}

%% file: 1_introduction.tex
\section{Introduction}
\label{sec:intro}
Since the introduction of Variational Auto-Encoders (VAEs)~\citep{kingma2013auto, rezende2014stochastic}, these so-called Deep Generative Models (DGMs) have established themselves as a go-to tool for learning representations of heterogeneous data sets. Their applications span financial time-series analysis~\citep{bergeron2022variational}, speech analysis and synthesis~\citep{9638604}, as well as biological data analysis~\citep{lopez2020enhancing}. Their natural ability to deal with multi-modal~\citep{wu2018multimodal}, temporal~\citep{9638604} and spatial data sets~\citep{yuan2019variational} makes them a powerful framework for extracting informative representations of data at a massive scale. Learning informative and compact representations from data is a milestone for applications driven by goodness of fit, or specific downstream prediction tasks. For many other cases, however, learning representations that are modular and have semantic meaning is crucial for reasons of interpretability. 

\begin{figure}[t]
    \centering
    \vspace{-10mm}
    \includegraphics[width=1\textwidth]{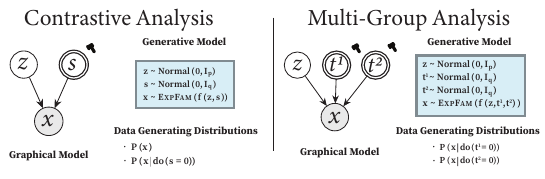}
    \vspace{-3mm}
    \caption{Presentation of the comparative deep generative models considered in this work.}
    \label{fig:generative-models}
    \vspace{-3mm}
\end{figure}

This paper is concerned with the problem of \textit{comparative analysis}, which seeks to model the similarities and differences across multiple data sets. This analytical approach is often driven by scientific applications, where researchers routinely juxtapose observations from a condition of interest (e.g., a disease) with a control condition (e.g., healthy). 

A particular form of comparative analysis is termed \textit{contrastive analysis}\footnote{This is a distinct line of work from the field of contrastive learning, that aims at distinguishing between positive and negative pairs of data points, as used in self-supervised learning.}. This methodology aims to characterize how a target data set differs from a background data set~\citep{zou2013contrastive} (Figure~\ref{fig:generative-models}, left). In achieving this, a generative model that incorporates two sets of latent variable $(\z, \s) \in \R^{p} \times \R^{q}$ is learned from data~\citep{abid2019contrastive, jones2021contrastive}. Here, the \textit{background variables} $\z$ represent patterns inherent to the background data set, while the \textit{salient variables} $\s$ capture nuances unique to the target data set. Contrastive analysis methods have been widely adopted in scientific research, with notable applications in omics data analysis~\citep{boileau2020exploring, weinberger2023isolating} and brain imaging studies~\citep{louiset2023sepvae}. Another approach within comparative analysis involves learning both shared and group-specific data representations using a single generative model~\citep{davison2019cross, weinberger22disentangling}. We refer to it as the multi-group analysis setting~(Figure~\ref{fig:generative-models}, right). 

The widespread success and adoption of these methods is somewhat surprising. Indeed, DGMs face important challenges in learning interpretable representations~\citep{locatello2019challenging}, and usually require ad-hoc regularization schemes in order to yield satisfactory performance~\citep{higgins2016beta, kim2018disentangling,lopez2018information}. Similarly, successful inference of comparative DGMs also requires the engineering of regularization approaches~\citep{weinberger2022moment}. This brings up the important theoretical question of why such regularization strategies are necessary. A plausible hypothesis is that the model itself is not identifiable, and that regularization helps constrain the function class used to fit the model. Here, non-identifiability means that given some data and a ground truth generating process, there exists an alternative model that has equal data likelihood, but such that the subspaces recovered from it are different. We note that identifiability is also an important question of its own, because it is a necessary condition to interpret and attribute semantic meaning to the learned representations, which is the end goal of real-world scientific applications.

We therefore explore the question of identifiability of comparative DGMs. To the best of our knowledge, such theoretical developments have been completely unexplored in the related literature. As a starting point, we highlight that data sets from different sources may be interpreted as the result of a do intervention on the graphical model~\citep{pearl2009causal}. This allows us to build upon recent advances in causal representation learning and non-linear ICA theory~\citep{khemakhem2020variational,lachapelle2022disentanglement,kivva2022identifiability} to prove the (block-wise) identifiability of comparative DGMs under the assumption of a piece-wise affine mixing function (e.g., parameterized by a ReLU / Leaky ReLU neural network). This demonstrates, for the first time, the identifiability of many recently published contrastive DGMs~\citep{jones2021contrastive, severson2019unsupervised, weinberger2023isolating}, and multi-group DGMs~\citep{severson2019unsupervised,weinberger22disentangling}. We also provide empirical evidence of this point with numerical experiments. This result is surprising, because of the practical need for regularization. To reconcile this apparent contradiction, we illustrate that identifiability guarantees are lost when the numbers of latent variables in each block are misspecified, as is often the case in real-world data analysis. In numerical experiments, we assess that existing regularization strategies considerably help mitigate this effect. Finally, motivated by this theoretical analysis, we also propose a new methodology for fitting comparative DGMs based on recent advances in multi-objective optimization~\citep{sener2018multi}, and constrained optimization~\citep{gallego2022controlled}. 

After briefly introducing the background (Section~\ref{sec:background}), we present our novel theory of identifiability for comparative DGMs (Section~\ref{sec:identifiability}). We then discuss limitations of the theory in the case of model misspecification in Section~\ref{sec:misspecification}. We propose novel algorithmic methodology (Section~\ref{sec:algo}), and conduct numerical experiments on simulations as well as a recent data set from a genetic screen profiled via single-cell RNA sequencing (Section~\ref{sec:experiments}). Discussion of related works appears in Appendix~\ref{app:related-work}.

%% file: 2_background.tex
\section{Background}
\label{sec:background}

This paper is concerned with the modular recovery of latent variables of a comparative analysis model that initially generated the data  (Figure~\ref{fig:generative-models}). Therefore, we briefly introduce the field of comparative analysis, and then present recent results on the identifiability of non-linear Independent Component Analysis (ICA), one of the prominent methods for latent variable recovery.

\subsection{Comparative Analysis with Deep Generative Models}
\paragraph{Contrastive Analysis} \cite{zou2013contrastive} introduced the goal of contrastive analysis, and the first algorithmic approaches (e.g, based on mixture models). \cite{abid2018exploring} proposed a contrastive principal component analysis (cPCA) method that captured intriguing variations from the target data set that do not appear in the background data set. Subsequent endeavors \citep{li2020probabilistic,jones2021contrastive} steered towards the creation of probabilistic latent variable models tailored to contrastive analysis, with a recent focus on deep generative models~\citep{severson2019unsupervised,abid2019contrastive,ruiz2019learning,weinberger2023isolating}. In this setting, a contrastive DGM has two sets of latent variables (Figure~\ref{fig:generative-models}, left): the salient variable $\s \in \R^p$ and the background latent variable $\z \in \R^q$. The \emph{target} data set is generated by sampling both sets of latent variables from an isotropic Gaussian prior distribution, passing them through a mixing function $f$ and sampling the data $\x$ from the exponential family distribution \textsc{ExpFam}, using $f(\z, \s)$ as the parameter. The \emph{background} data set is generated similarly, but by setting $\s = \zero$ to make sure that $\s$ is utilized only for describing the target data set. We denote this distribution as a hard intervention $p_\theta\left(\x \mid \text{do}(\s=\bm{0})\right)$~\citep{pearl2009causal} in order to draw parallels with interventional causal representation learning~\citep{ahuja2023interventional}.

In terms of inference procedure, all methods rely on the contrastive Variational Auto-Encoder (cVAE) framework~\citep{abid2019contrastive}. For each sample in the target data set (resp. the background data set), the variational distribution is $q_\phi(\z \mid \x)q_\phi(\s \mid \x)$ (resp. $q_\phi(\z \mid \x)$ only, as $\s=\zero$). Then, the composite evidence lower bound (ELBO) is derived as
\begin{align}
\label{eq:elbo}
    \E_{q_\phi(\z \mid \x)}\log \frac{p_\theta(\x, \z, \zero)}{q_\phi(\z \mid \x)} + \E_{q_\phi(\z \mid \x)q_\phi(\s \mid \x)}\log \frac{p_\theta(\x, \z, \s)}{q_\phi(\z \mid \x)q_\phi(\s \mid \x)}.
\end{align}
This composite ELBO corresponds to the sum of two individual ELBOs for each of the two data sets (background and target). It may be used as an objective function for maximization, in conjunction with adequate regularization of the neural networks parameterizing the variational distribution as a function of the input data. In~\citet{abid2019contrastive}, an additional regularization term specifically promotes independence of the two sets of latent variables. 

\paragraph{Multi-group Analysis} Recently, several models have been designed to distinguish patterns that are shared by all data sets, versus the ones that are specific to each data set~\citep{davison2019cross,weinberger22disentangling}. We provide more details about the generative model and the inference mechanism in Appendix~\ref{app:multi-group}. 

\subsection{Identifiability of Non-linear Independent Component Analysis}

ICA assumes that $\x \in \mathbb{R}^d$ is generated using $p$ independent latent variables $\z = (z_1, \ldots, z_p)$, called \emph{independent components} \citep{Hyvarinen2002independent}. More precisely, observations $\x$ are defined as $\x = f(\z) + \bm{\epsilon}$ with $f$ a mixing function and $\bm{\epsilon}$ an exogenous noise variable. The ICA literature established that in the general case of a non-linear \emph{mixing function} $f$, the model is unidentifiable from i.i.d. observations of $\x$~\citep{hyvarinen1999nonlinear}, and therefore the original $\z$ may not be recovered. Given this negative result, several papers introduced identifiable forms of non-linear ICA models~\citep{harmeling2003kernel, sprekeler2014extension, hyvarinen2016unsupervised, hyvarinen2017nonlinear}, based on the observability of an \emph{additional auxiliary} random variable. However, such auxiliary variable is not always available in practice. More recently, ~\cite{kivva2022identifiability} proposed a new theory of identifiability based on the assumption that $f$ is a piece-wise affine function. Their main result is that many previously proposed deep generative models parameterized with ReLU / Leaky ReLU neural networks and additive Gaussian observation noise, a commonly used architecture, are identifiable up to a linear transformation of the mixing function. Our work directly builds upon this line of work to assess the identifiability of comparative DGMs.

%% file: 3_identifiability_theory.tex
\section{A Theory of Identifiability for Comparative Analysis Models}
\label{sec:identifiability}
For the sake of conciseness and ease of notation, we focus on the contrastive analysis case in this section. Definitions, theorem statements, and proofs for the multi-group setting appear in Appendix~\ref{app:multi-group-proofs}.

\subsection{Subspace Identifiability}
Identifiability is a critical property to understand whether a model's parameters can be uniquely inferred from observations~\citep{ran2017parameter}. Within the context of contrastive analysis, our concern is not the recovery of latent variables at the component level. Rather, our interest lies in the retrieval of specific subspaces, namely the blocks $\mathcal{Z} = \R^p$ and $\mathcal{S} = \R^q$. To introduce this concept, we first define a general criterion for compatibility of the Cartesian product of subspaces by a map.
\begin{definition}[Compatible map]
    Let $(E_1, \ldots, E_n)$ be $n$ Euclidean spaces, and let $E = \prod_{i=1}^nE_n$ designate the Cartesian product space. A map $\phi: E \rightarrow E$ is said to be \emph{compatible} with the Cartesian product $E = \prod_{i=1}^nE_n$ if there exist maps $\left(\bar{\phi}_1, \ldots,\bar{\phi}_n\right)$ of the subspaces $(E_1, \ldots, E_n)$ such that for all $\bm{e} = (e_1, \ldots, e_n) \in E$, we have that $\phi(\bm{e}) = \left(\bar{\phi}_1(e_1), \ldots, \bar{\phi}_n(e_n)\right)$.
\end{definition}
Now, if we denote the support of latent variables for the background data set (resp. the target data set) as $\mathcal{D}^b = \mathcal{Z} \times \{0\}$ (resp. $\mathcal{D}^t =\mathcal{Z} \times \mathcal{S}$), we define the subspace disentanglement condition as follows.
\begin{definition}[Subspace Disentanglement]
Let $f$ be the ground truth mixing function, and $\tilde{f}$ be a learned mixing function. Let us also assume that $f(\mathcal{D}^b) = \tilde{f}(\mathcal{D}^b)$,  $f(\mathcal{D}^t) = \tilde{f}(\mathcal{D}^t)$, and that the map $v = f^{-1} \circ \tilde{f}$ is well-defined. $\tilde{f}$ is said to be subspace-disentangled with respect to $f$ if $v$ is compatible with respect to the Cartesian product $\mathcal{D}^t =\mathcal{Z} \times \mathcal{S}$.
\end{definition}
When this property is not verified, the learned mixing function $\tilde{f}$ and the background $f$ provide distinct decompositions of the signal from the feature space $\mathcal{X}$ into $\mathcal{Z}$ and $\mathcal{S}$. Related definitions that appear in previous works such as~\citet{von2021self} are discussed in Appendix~\ref{app:related-work}. We may now outline our definition for subspace identifiability of contrastive DGMs.
\begin{definition}[Subspace Identifiability]
A contrastive analysis model with ground truth mixing function $f$ is subspace identifiable from data if for all other mixing functions $\tilde{f}$ that yield the same background and target data distributions, we have that $\tilde{f}$ is subspace-disentangled with respect to $f$. 
\end{definition}
When such a model is subspace identifiable, and we observe data $\x$, we are guaranteed that the learned representations $(\tilde{\z}, \tilde{\s}) = \tilde{f}^{-1}(\x)$ are given by a transformation of each of the original spaces: $\tilde{\z} = h_{\z}(\z)$, and $\tilde{s} = h_{\s}(\s)$, and therefore semantic meaning can be attributed to those subspaces. 

Although the fact that we observe two data sets is potentially helpful in breaking symmetry in the roles played by the shared latent variables $\z$ and $\s$, it is not true in general that all contrastive analysis models are subspace identifiable.
\begin{example}[Counterexample]
    For $p=2$ and $q=1$, let us consider the following map of $\R^3$:
\begin{equation}
\Phi: \left(\begin{array}{l}
z_1 \\ z_2 \\ s
\end{array}
\right) \mapsto
\left(\begin{array}{c}
z_1 \cos s - z_2 \sin s \\ z_1 \sin s + z_2 \cos s \\ s
\end{array}
\right).
\end{equation}
\end{example}
For any non-trivial mixing function $f$, we define $\tilde{f} = f \circ \Phi$. $\tilde{f}$ and $f$ generate the same data distributions because $\Phi$ is a diffeomorphism that preserves volume, and distance to the origin. However, $\tilde{f}$ is not subspace disentangled with respect to $f$. The complete proof appears in Appendix~\ref{app:general-counterexample}. Example~1 may be seen as an extension of the classical counter-example of identifiability for linear ICA~\citep{Hyvarinen2002independent}, exploiting the rotational invariance of the Gaussian distribution but with a non-constant rotation angle. 

\subsection{Identifiability Result for Piece-wise Affine Mixing Functions and Noiseless Observations}

The counterexample presented above suggests that we must restrict the function class for $f$ in order to potentially obtain identifiability. We propose to build upon recent work on identifiability of DGMs with mixing functions specified as multilayer perceptrons (MLP) with ReLU / Leaky ReLU activations~\citep{kivva2022identifiability} to obtain the first result of identifiability of comparative analysis models. 
\begin{restatable}[Identifiability Theorem]{theo}{theorelu}
    \label{theo:relu}
    Let the ground truth mixing function $f$ and the learned mixing function $\tilde{f}$ both be continuous and injective piece-wise affine mixing functions such that $f(\z, \s) \equald \tilde{f}(\z, \s)$ and $f(\z, \zero) \equald \tilde{f}(\z, \zero)$. Then, $\tilde{f}$ is subspace disentangled with respect to $f$ and the noiseless version of the contrastive analysis model is subspace identifiable.
\end{restatable}
The proof appears in Appendix~\ref{app:relu}, and consists of two steps. First, we apply the result of~\citet{kivva2022identifiability} to each of the target and background data distributions to obtain the linear identifiability of the mixing function on each domain. Then, we rely on the geometry of affine transformations to prove that the disentanglement criterion must hold on both data domains. Because this is an instance of linear disentanglement, this implies that $v$ is a linear transformation. We also note that the assumptions of isotropic Gaussian distributions for $p(\z)$ and $p(\z, \s)$ for Theorem~\ref{theo:relu} could be relaxed to members of an exponential family of distributions, as long as the densities are analytic functions, and the family is closed under additive transformation \citep{kivva2022identifiability}.

Because the problem of identifiability of non-linear ICA with additive Gaussian noise can be reduced to the noiseless case~\citep{khemakhem2020variational}, Theorem~\ref{theo:relu} and its multigroup variant, Theorem~\ref{theo:relu-group}, are readily applicable to several real-world models. In the special case where $f$ is linear injective, these results yield the identifiability of probabilistic contrastive principal component analysis \citep{li2020probabilistic}, and multi-study factor analysis \citep{devito2019multi}. The non-linear version provides the identifiability of the cross-population VAE~\citep{davison2019cross}, and of the contrastive VAE (cVAE)~\citep{abid2018exploring}.

\subsection{Extensions towards models with Observational Count Noise}

The results from Theorem~\ref{theo:relu}, and to the best of our knowledge, all previous results on identifiability of non-linear ICA models\footnote{The initial version of \citet{khemakhem2020variational} presented a proof of identifiability for categorical variables that has since been removed due to a mistake in the write-up.}
only apply to noiseless measurements, or to Gaussian observation noise. However, many real-world applications of DGMs~\citep{lopez2020enhancing}, and especially comparative DGMs, have been proposed to deal with count data, such as the contrastive generalized latent variable model (CGLVM)~\citep{jones2021contrastive}, ContrastiveVI~\citep{weinberger2023isolating} and multiGroupVI~\citep{weinberger22disentangling}. We therefore now show that the non-linear ICA identifiability problem with Poisson or negative binomial noise reduces to the noiseless one. 

\begin{restatable}[Reduction from observational count noise to the noiseless setting]{theo}{theocounting}
\label{theo:counting}
    \\ Let $\uvar \sim \textrm{Normal}\left(0, I_p\right)$. Let $f = \sigma \circ g$ (resp. $\tilde{f} = \sigma \circ \tilde{g}$) be the composition of a scalar link function $\sigma$, valued in $\R_+$ (applied component-wise), with a piecewise affine function $g$ (resp. $\tilde{g}$). Let $\x \sim p_x(f(\uvar))$ and $\tilde{\x} \sim p_x(\tilde{f}(\uvar))$ such that $p_x$ is Poisson or negative binomial with fixed shape. If $\sigma$ is a bicontinuous bijection, then, 
    \begin{align}
        \tilde{\x} \equald \x \implies \tilde{f}(\uvar) \equald f(\uvar) \implies \tilde{g}(\uvar) \equald g(\uvar).
        \end{align}
\end{restatable}
The proof appears in Appendix~\ref{app:poissonaffine}. Our proof appeals to calculation and identification of the Laplace transformation of the distribution of random variables $\x$ and $\tilde{\x}$. We note that more general versions of this theorem were introduced in early identifiability theory~\citep{Sapatinas1995IdentifiabilityOM,Teicher1961IdentifiabilityOM}. From Theorem~\ref{theo:counting}, we conclude to the block-identifiability of the comparative analysis models mentioned above in the setting of observational count noise and invertible link function.

Additionally, we prove that identifiability does not hold in the case of Bernoulli observational noise without further assumptions. Explicit counterexamples appear in Appendix~\ref{app:ber}, disproving a conjecture in~\citet{khemakhem2020variational}. We instead hypothesize that non-identifiability holds in general, for any observational distribution with fixed finite support.

%% file: 4_impact_regularization.tex
\section{Impact of Misspecification}
\label{sec:misspecification}
Our main results (Theorems~\ref{theo:relu} and~\ref{theo:counting}) implicitly assume that the observed data have been simulated from the generative model $p_\theta(\x)$. However, this may be impossible to verify in practice, as there are many assumptions that might be unknown to practitioners. Examples of such assumptions include the specification of the graphical model, a function class for the mixing function $f$, as well as the number of latent variables. Given any source of such model misspecification, the theory above unfortunately does not apply. 

We focus in this work on a discussion of the impact of a misspecification of the number of latent variables. This is an important starting point, because it is easy to illustrate, and it is known that overestimating the number of latent variables induces severe entanglement in practice, making regularization necessary~\citep{weinberger2022moment}. Beside empirical work, theoretical developments are needed to understand how this occurs in the contrastive analysis setting.

To illustrate this, let $p' \geq p$ and $q' \geq q$ be the estimated dimensions of the background space and salient space, respectively, with $p' + q' > p + q$. Further, denote by $\tilde{\z} = (\z, \uvar)$ and $\tilde{\s} = (\s , \vv)$ the respective latent variables, where $\uvar$ and $\vv$ are additional variables of dimensions $p'-p$ and $q'-q$, respectively. We consider data $\tilde{\x} = \tilde{f}(\tilde{\z}, \tilde{\s})$, generated by a learned mixing function $\tilde{f}$. Compared to our previous setting, we cannot assume injectivity of $\tilde{f}$ under equality of the data generating distributions. Indeed, if we assume that $\tilde{f}(\tilde{\z}, \tilde{\s}) \equald f(\z, \s)$, then the support of those distributions must be equal $\tilde{f}(\R^{p' + q'}) = f(\R^{p + q})$ and have the same manifold dimension. But because of the dimension mismatch, $\tilde{f}$ cannot be injective. In particular, the lack of injectivity of $\tilde{f}$ implies that it does not have a well defined inverse, and makes theoretical analysis challenging.

We therefore first seek to characterize the case where both $f$ and $\tilde{f}$ are linear functions. Surprisingly perhaps, we show that entanglement does not occur in this scenario.
\begin{restatable}[Block-wise identifiability under misspecification for the linear case]{prop}{propidentlinear}
\label{prop:linidentmisspec}
\\Let the ground truth mixing function $f$ be injective linear, and the learned function $\tilde{f}$ be a linear function such that $f(\z, \s) \equald \tilde{f}(\tilde{\z}, \tilde{\s})$ and $f(\z, \zero) \equald \tilde{f}(\tilde{\z}, \zero)$. Then, there exist surjective linear functions $h_{\z}$ and $h_{\s}$ such that $(\z, \s) = v(\tilde{\s}, \tilde{\z}) = \left(v_{\s}(\tilde{\s}), v_{\z}(\tilde{\z})\right)$, where $v = f^{-1} \circ \tilde{f}$. 
\end{restatable}
The proof appears in Appendix~\ref{app:identif_misspec_linear_model} and builds upon the proof of identifiability for factor analysis. This result is interesting, as it may explain why regularization is not used for linear comparative analysis models, but was introduced with the first applications of DGMs to this setting~\citep{abid2019contrastive}. We introduce a broad class of examples of non-identifiable models with non-linear and non-injective mixing functions in Appendix~\ref{app:nonlinmisspec}.

Although the discussion above is important to define what the lack of identifiability could imply, it ignores the impact of the variational inference procedure. This is a central point, because the regularization approaches introduced for comparative DGMs impose independence constraints for the aggregated variational posterior~\citep{salakhutdinov2010efficient}. For the target data set, the aggregated posterior is defined as \(
    \hat{q}_\phi^t(\tilde{\z}, \tilde{\s}) = \E_{p_\textrm{data}(\x)}\left[q_\phi(\tilde{\z} \mid \x)q_\phi(\tilde{\s} \mid \x)\right],
\)
and the regularizer aims to enforce the independence statement $\hat{q}_\phi^t(\tilde{\z}) \indep \hat{q}_\phi^t(\tilde{\s})$~\citep{abid2019contrastive}. Other regularization approaches are described in Appendix~\ref{app:regularization}. 

Interestingly, the independence constraint may not be enough to restore identifiability in general. Our conjecture is that it does contribute to reducing entanglement by constraining the inference network, and therefore restricting the space of admissible mixing functions. We demonstrate in later sections empirical evidence that it indeed improves disentanglement, but leave theoretical analysis to future work.

%% file: 5_algorithm.tex
\section{Multi-Objective Constrained Optimization for Contrastive VAEs (MO-CO-cVAEs)}
\label{sec:algo}
The standard routine for fitting comparative DGMs consists in casting the inference problem as an optimization problem by maximizing a lower bound on the likelihood, following the principles of variational inference~\citep{jordan1999introduction}. By more closely inspecting the nature of the optimization problem at hand, we present here a novel method for fitting contrastive DGMs. 

\subsection{Maximum Likelihood Across Data Sets Using Multi-Objective Optimization}
Existing methodology for fitting comparative analysis models typically derives one evidence lower bound (ELBO) for each data set, specifically, $\mathcal{L}^B({\theta, \phi})$ for the background $\mathcal{L}^T({\theta, \phi})$ for the target data set. The objective function is then defined as the sum of these ELBOs 
as in Equation~\ref{eq:elbo}. We refer to this approach as the Single Objective cVAE (SO-cVAE). In this scenario, optimizing one loss may negatively impact the optimization of the other, a common challenge in multi-task learning~\citep{sener2018multi}. Moreover, our theoretical insights indicate that the learned parameters for the generative model should be optimal across all considered data sets. As a result, we advocate for framing this problem of inference across multiple data sets as a multi-objective optimization problem:
\(
    \min_{\theta, \phi} \left(-\mathcal{L}^B({\theta, \phi}),- \mathcal{L}^T({\theta, \phi})\right),
\)
that we solve using the Multiple-Gradient Descent Algorithm~\citep{desideri2012multiple}, where at each step $t$, the direction $\delta_t$ used for the descent is a convex combination of the gradient of each ELBO:
\begin{align}
    \delta_t &= -\alpha_t\nabla\mathcal{L}^B({\theta^t, \phi^t}) - (1-\alpha_t)\nabla\mathcal{L}^T({\theta^t, \phi^t})\\
    \alpha_t &= \argmin_{\alpha \in [0, 1]} \norm{\alpha\nabla\mathcal{L}^B({\theta^t, \phi^t}) + (1-\alpha)\nabla\mathcal{L}^T({\theta^t, \phi^t})}_2^2, \label{eq:alpha}
\end{align}
where the quadratic optimization problem in Equation~\ref{eq:alpha} admits a closed-form solution (Appendix~\ref{app:algo}). This procedure provably converges to a Pareto-optimal design point in the batch setting (\citet{zhou2022on} discusses the stochastic setting). In our implementation, we solely rely on gradients of the last layer of the decoder to approximately calculate the optimal weight $\alpha$, and have observed satisfactory performance. We refer to this approach as the Multiple Objective cVAE (MO-cVAE).

\subsection{Interpretable Hyperparameter Selection via Constrained Optimization}
The most common approach to regularize models in the comparative analysis literature involves adding a penalization term to the ELBO, leading to solving an unconstrained optimization problem (U-cVAE). For example, independence constraints are typically enforced via penalization of mutual information approximations, estimated either via the density-ratio trick~\citep{abid2019contrastive}, or kernel-based embedding of distributions~\citep{weinberger2022moment}. This approach has two significant drawbacks. First, it necessitates the calibration of the Lagrangian multiplier. Currently, the only methods available involve either using a preset value~\citep{abid2019contrastive} or comparing the scales of loss functions~\citep{weinberger22disentangling}. Second, the primary goal of the optimization procedure is not to minimize the mutual information between the latent variables, but for the mutual information to be \textit{sufficiently low} for practical considerations. 

For these reasons, we instead explore the design of a constrained optimization problem that enhances interpretability and automation for the selection of the Lagrangian parameter:
\begin{align}
    \min_{\theta, \phi} \mathcal{L}({\theta, \phi}) = -\mathcal{L}^B({\theta, \phi}) - \mathcal{L}^T({\theta, \phi}) \text{~~such that~~} \text{CKA}\left(\hat{q}_\phi(\z, \s)\right) \leq \beta,
\end{align}
where $\beta > 0$ is a scalar 
, and $\text{CKA}\left(\hat{q}_\phi(\z, \s)\right)$ is the centered kernel alignment metric~\citep{kornblith2019similarity}, a non-parametric measure of correlation between random vectors. Given two positive definite kernels $k: \R^p \times \R^p \rightarrow \R$, and $l: \R^q \times \R^q \rightarrow \R$, we may define the cross-covariance operator $C_{\z, \s}$ that embeds the joint distribution $\hat{q}_\phi(\z, \s)$ as a linear operator in the RKHS obtained from both kernels~\citep{gretton12akernel}. Similarly, we may embed each marginal distribution $\hat{q}_\phi(\z)$ and $\hat{q}_\phi(\s)$ as linear operators $C_{\z, \z}$ and $C_{\s, \s}$. Then, the CKA is defined as:
\begin{align}
    \text{CKA}\left(\hat{q}_\phi(\z, \s)\right) = \frac{\norm{C_{\z, \s}}_{\text{HS}}^2}{\norm{C_{\z, \z}}_{\text{HS}}\norm{C_{\s, \s}}_{\text{HS}}},
\end{align}
where $\norm{.}_{\text{HS}}$ designates the Hilbert-Schmidt norm of a linear operator in the RKHS. When the kernels are linear, the CKA metric becomes related to the RV-coefficient~\citep{robert1976unifying} as well as Tucker's congruence coefficient~\citep{tucker1951method}, both practically used to estimate correlations between pairs of random vectors. Throughout this manuscript, we use a maximal CKA value of $\beta = 0.05$ to obtain satisfactory performance. To solve the constrained optimization problem, we consider the following equivalent Lagrangian (details appear in Appendix~\ref{app:optim-constrained}),
\begin{align}
    \min_{\theta, \phi} \max_{\lambda \geq 0} \mathcal{L}^\lambda({\theta, \phi}) = \mathcal{L}({\theta, \phi}) + \lambda \left(\norm{C_{\z, \s}}_{\text{HS}}^2 - \beta\norm{C_{\z, \z}}_{\text{HS}}\norm{C_{\s, \s}}_{\text{HS}}\right).
\end{align}
We perform simultaneous gradient descent on $(\theta, \phi)$ and projected gradient ascent on the Lagrangian $\lambda$ associated with the constraint~\citep{lin2020gradient}:
\begin{align}
    [\theta^{t+1}, \phi^{t+1}] &= [\theta^{t}, \phi^{t}] - \eta_{\text{primal}} \nabla\mathcal{L}^\lambda(\theta^t, \phi^t) \label{eq:primal}\\
    \lambda^{t+1} &= \left[\lambda^t + \eta_{\text{dual}} \left(\norm{C_{\z, \s}}_{\text{HS}}^2 - \beta\norm{C_{\z, \z}}_{\text{HS}}\norm{C_{\s, \s}}_{\text{HS}}\right)\right]_{+}\label{eq:dual},
\end{align}
where $[a]_+ = \max(0, a)$. We obtain a stochastic estimate of the gradient $\nabla\mathcal{L}^\lambda(\theta^t, \phi^t)$ by subsampling data points, as well as latent variables from the variational distribution. We estimate the Hilbert-Schmidt norms using the Hilbert-Schmidt Independence Criterion~\citep{gretton12akernel}. More precisely, from samples $(z_i, s_i)_{i=1}^M$ from $\hat{q}_\phi(\z, \s)$, the kernel matrices $K_z$ (and $K_s$) are defined as $K_{ij} = k(z_i, z_j)$, and the HSIC is defined as $\text{HSIC}(K, L) =  (M - 1)^{-2} \text{Tr}(KHLH)$, where $H = I - \frac{1}{M}\bm{11}^\top$ is a centering matrix. Then, $\text{HSIC}\left(K_{z}, K_s\right)$ is an unbiased estimator for $\norm{C_{\z, \s}}_{\text{HS}}^2$, and we proceed similarly for the remaining terms~\citep{gretton12akernel}.

In our experiments, we use the Adam optimizer~\citep{kingma2014adam} for the gradient step described in Equation~\ref{eq:primal}, with a learning rate of $\eta_{\text{primal}} = 0.001$ and gradient ascent with a learning rate of $\eta_{\text{dual}} = 1$ for updating the Lagrangian coefficient in Equation~\ref{eq:dual}. We refer to this method as the COnstrained cVAE (CO-cVAE). The reader will notice that this approach may be combined with the multi-objective optimization approach described above, in which case we refer to it as MO-CO-cVAE.

%% file: 6_experiments.tex
\section{Experiments}
\label{sec:experiments}
We present empirical evidence of our theory of identifiability with a simulation framework, where data is generated with a piece-wise linear mixing function. Then, we apply our proposed optimization framework to a real-world example from single-cell perturbation data analysis. We base our experiments upon the implementation of contrastive DGMs presented in~\cite{weinberger2023isolating}. All results are reported with mean and standard deviation over $5$ random initializations. Additional experimental results and supplementary metrics appear in Appendix~\ref{app:results_supp}.

\subsection{Synthetic Data Experiments}

In order to provide empirical evidence for our theory of identifiability, we generated synthetic data in the contrastive analysis framework, according to the generative model described in Figure~\ref{fig:generative-models}, where $f$ is a four-layers Leaky ReLU neural network (details in Appendix~\ref{app:simulation}).

\paragraph{Verification of Identifiability} Our theory dictates that if the number of latent variables in each space is known, then the contrastive model is identifiable. In addition, we know that the latent variables should be linear transformations of each other, with no leakage between the distinct spaces. To verify this claim, we generated data with $p=q=5$, and fitted a contrastive DGM with the same estimated number of latent dimensions (no regularization). We refer to those \textit{unregularized} models as MO-cVAE and SO-cVAE, depending on whether the multi-objective optimization procedure was applied or not. We quantified the level of disentanglement using the Pearson Mean Correlation Coefficient after a linear transformation (MCC) between ground truth latent variables, and estimated latent variables (Appendix~\ref{app:metrics}). More specifically, we calculated the Pearson MCC between $\hat{\z}$ and $\z$, $\hat{\s}$ and $\s$ (higher is better), but also between $\hat{\z}$ and $\s$, and $\hat{\s}$ and $\z$ (lower is better). As an aggregated metric, we define the $\delta$-MCC $\in [0, 1]$ as~\useshortskip
\begin{align}
    \delta\text{-MCC} = \frac{1}{2}\left(\text{MCC}_{\hat{\z}\z} + \text{MCC}_{\hat{\s}\s}\right) + \frac{1}{2}\left(\text{MCC}_{\hat{\s}\z} - \text{MCC}_{\hat{\z}\s}\right).
\end{align}\useshortskip
The results highlight high conservation of each individual latent space, and remarkably low leakage between latent spaces, for both Poisson, and negative binomial observation models~(Table~\ref{tab:simu_mcc}). For reference, we also fitted a vanilla VAE, with the same architecture and noise model. Because the VAE only yields one set of latent variables, we ran the method with $p + q$ number of latent variables and used contrastive PCA to split the latent space into  the background or the salient space (Appendix~\ref{app:baselines}). We observe poor performance of the VAE for this task, likely because it treats all samples as independent and identically distributed, and ignores additional knowledge about the background data set. This suggests, as expected, that the additional assumptions enforced by contrastive DGMs are necessary for identifiability. 

\input{tables/simu_mcc}

\paragraph{Impact of Misspecification} Then, we wanted to illustrate the fact that disentanglement performance drops in the setting of misspecification of the number of latent variables. Towards this goal, we maintained $p$ and $q$ to $5$ in the simulation framework, but augmented $\hat{q}$, the dimensionality of $\s$ in the inference method. 
We noticed a strong degradation in the performance, as highlighted in the drop in $\delta$-MCC (Table~\ref{tab:mispec}). Careful examination of the individual MCC scores revealed a high leakage from the ground truth background variables $\z$ into the estimated salient variables $\hat{\s}$~(Table~\ref{tab:mispec_app} and~\ref{tab:mispec_app_spearman}). 

\paragraph{Information Constraints Improve Performance} We then explored how much the independence constraints could help improve the performance in the case $\hat{q}=10$. Towards this end, we applied the HSIC penalty with a fixed scaling factor $\lambda$, denoted as U-SO-cVAE and U-MO-cVAE (U stands for unconstrained), as well as the constrained optimization procedure with $\beta = 0.05$, denoted as CO-SO-sVAE and CO-MO-cVAE. We report values of the $\delta$-MCC for different values of the regularization strength in Table~\ref{tab:mispec-reg}. Although regularization helped partially restore the performance in some sensible range of $\lambda$, the constrained optimization approach is more practical as $\lambda$ is adjusted automatically during training, and achieved competitive performance. 

\paragraph{Comparison with other Regularizers} In order to justify that the HSIC penalty is a competitive regularizer, we also assessed the performance of the regularization from both ConstrastiveVI~\citep{weinberger2023isolating}, as well as the original cVAE~\citep{abid2019contrastive} on the same benchmark in the experiment from Table~\ref{tab:mispec-reg}. The ContrastiveVI regularization seems to help ($\delta$-MCC value of 0.75), but its performance remains lower than CO-MO-cVAE. cVAE achieves a poorer result ($\delta$-MCC value of 0.69), slightly improving over the unregularized method. 

\paragraph{Multi-objective Optimization } We also note that the approach that used multi-objective optimization systematically performed better throughout this benchmark (Tables~\ref{tab:simu_mcc}, ~\ref{tab:mispec}, and~\ref{tab:mispec-reg}). 

\input{tables/simu_regularization}

\subsection{Single-cell Perturbation Analysis}

As an application to real data, we present an experiment focused on characterizing the effect of genetic perturbations on single-cell gene expression levels, a central problem in modern molecular biology~\citep{dixit2016perturb, norman2019exploring}. In these data sets, we observe two important sources of variation. First, cells react to the genetic perturbation they were exposed to, and modulate the expression level of their genes. Second, there is some inherent variation in gene expression levels that happens due to the cells going through biological processes such as stages of the cell cycle, or simply due to heterogeneity in the initial population of cells. An important problem therefore consists in disentangling these effects, and ContrastiveVI~\citep{weinberger2023isolating} was conceived with this goal in mind. 

We focus on a recent data set~\citep{norman2019exploring} that contains expression profiles from $33,820$
erythroleukemia (cancer) cells, after interventions targeting each of $105$ genes, as well as $131$ pairs of those same genes. Each measurement from a single-cell combines the identity of the intervention (target genes) and a count vector where each entry
is the expression level of each gene in the genome. Because of experimental limitations~\citep{grun2014validation}, we
observe signal only for a subset of several thousand genes out of
the approximately $20,000$ genes in the genome.
Here we selected $d = 2,000$ genes. The goal
of the experiment was to manipulate gene pairs and measure the resulting changes in cell state to gain insights into how complex phenotypes emerge and identify genes that interact to promote differentiation to a specific cell state.

\input{tables/norman_main}

To assess the performance of each method, we first evaluated how well the salient space captures the effect of perturbations. Specifically, we clustered cells based on their salient embeddings and assessed how well those clusters overlap with known biological labels attributed to each of the perturbations, following~\citet{weinberger2023isolating}. We reported the Adjusted Rand Index (ARI), the Normalized Mutual Information (NMI), as well as the Average Silhouette Width (ASW). In addition, we assessed the overlap in content between the two latent spaces by training a linear regression model from one space to the other, and reporting the MCC (cMCC-P refers to the Pearson correlation and cMCC-S to the Spearman correlation).

We applied our evaluation pipeline for the VAE, MO-CO-cVAE, SO-CO-cVAE, as well as non-regularized variants that we note as MO-cVAE and SO-cVAE (Table~\ref{tab:real}). We observed again that the multi-objective variant outperforms the single-objective method, and that constrained optimization provided an effective regularization strategy. As a point of comparison, we also reported the performance of ContrastiveVI and the original cVAE, and noticed that MO-CO-cVAE performed better. This suggests that our novel methodology is effective in practice. Qualitative comparisons appear in Appendix~\ref{app:add-results}.

%% file: tables/simu_mcc.tex
\begin{table}[t]
\centering
\vspace{-5mm}
\caption{Identifiability under assumptions of known dimensions of latent spaces. Best in bold.}\vspace{2mm}
\scalebox{0.9}{
\begin{tabular}{lc|ccccc}
\toprule
     \textbf{Model}         &     \textbf{Noise}                                                                                  & \textbf{MCC}$_{\hat{\z}\z}$ ($\uparrow$) & \textbf{MCC}$_{\hat{\z}\s}$ ($\downarrow$) & \textbf{MCC}$_{\hat{\s}\z}$ ($\downarrow$) & \textbf{MCC}$_{\hat{\s}\s}$ ($\uparrow$) & $\delta$\textbf{-MCC} ($\uparrow$) \\\midrule
\textbf{MO-cVAE} & \multirow{3}{*}{\textbf{Poisson}}                                                     &   $0.91 \pm 0.01$           &   $\mathbf{0.08 \pm 0.01}$            &    $\mathbf{0.07 \pm 0.01}$            &    $\mathbf{0.94 \pm 0.01}$            &          $\mathbf{0.85 \pm 0.01}$     \\
\textbf{SO-cVAE} &                                                     &   $\mathbf{0.93 \pm 0.01}$           &   $0.13 \pm 0.01$            &    $\mathbf{0.07 \pm 0.02}$            &    $0.92 \pm 0.01$            &          $0.83 \pm 0.01$     \\
\textbf{VAE}  &  & $0.87 \pm 0.04$           &   $0.17 \pm 0.9$            &    $0.14 \pm 0.07$            &    $0.92 \pm 0.04$            &          $0.74 \pm 0.12$     \\ \midrule
\textbf{MO-cVAE} & \multirow{3}{*}{\textbf{\begin{tabular}[c]{@{}l@{}}Negative\\ binomial\end{tabular}}} &      $\mathbf{0.93 \pm 0.01}$           &   $0.10 \pm 0.01$            &    $\mathbf{0.06 \pm 0.01}$            &    $\mathbf{0.94 \pm 0.01}$            &          $\mathbf{0.83 \pm 0.01}$     \\
\textbf{SO-cVAE} &  &      $0.92 \pm 0.01$           &   $\mathbf{0.08 \pm 0.01}$            &    $0.07 \pm 0.01$            &    $\mathbf{0.94 \pm 0.01}$            &          $\mathbf{0.83 \pm 0.01}$     \\
\textbf{VAE}   & &      $0.81 \pm 0.10$           &   $0.46 \pm 0.18$            &    $0.37 \pm 0.18$            &    $0.80 \pm 0.10$            &          $0.39 \pm 0.28$     \\
\bottomrule
\label{tab:simu_mcc}
\end{tabular}
}
\vspace{-5mm}
\end{table}

%% file: tables/simu_regularization.tex
\begin{table}[ht]
\vspace{-5mm}
\begin{minipage}[t]{0.45\textwidth}
\centering
\caption{Impact of misspecification of latent dimensionality on $\delta$-MCC.\label{tab:mispec}}
\scalebox{1}{
\begin{tabular}{lcc}
\toprule
$\hat{q}$ & \textbf{SO-cVAE} & \textbf{MO-cVAE} \\ 
            \midrule
\textbf{5} &          $\mathbf{0.83 \pm 0.01}$  &          $\mathbf{0.85 \pm 0.01}$\\
\textbf{7}  & $0.73 \pm 0.03$ & $0.81 \pm 0.01$ \\
\textbf{10} & $0.66 \pm 0.01$  & $0.75 \pm 0.01$ \\
\textbf{15} & $0.58 \pm 0.02$ & $0.70 \pm 0.02$ \\ \bottomrule
\end{tabular}
\vspace{-3mm}
}
\end{minipage}\hfill
\begin{minipage}[t]{0.5\textwidth}
\centering
\caption{Impact of regularization on $\delta$-MCC under misspecification.\label{tab:mispec-reg}}
\vspace{-3mm}
\scalebox{1}{
\begin{tabular}{lcc}\toprule
Regularization& \textbf{SO-cVAE} & \textbf{MO-cVAE}\\
\midrule
\textbf{U} ($\lambda=0$) &  $0.66 \pm 0.01$ & $0.75 \pm 0.01$ \\
\textbf{U} ($\lambda = 10$) & $0.70 \pm 0.02$ &  $0.78 \pm 0.01$  \\
\textbf{U} ($\lambda = 50$) & $0.76 \pm 0.01$ & $0.79 \pm 0.02$\\
\textbf{U} ($\lambda = 100$) & $0.66 \pm 0.08$ & $0.80 \pm 0.01$ \\ 
\textbf{U} ($\lambda = 200$) & $0.32 \pm 0.10$ & $0.34 \pm 0.13$ \\ 
\textbf{CO}  & $\mathbf{0.77 \pm 0.01}$ & $\mathbf{0.80 \pm 0.01}$ \\ \bottomrule
\end{tabular}
}
\end{minipage}
\vspace{-3mm}
\end{table}

%% file: tables/norman_main.tex
\begin{table}
\centering
\vspace{-5mm}
\caption{Results on real data.}\vspace{2mm}
\label{tab:real}
\scalebox{1}{
\begin{tabular}{lccc|cc}
\toprule
 & \textbf{ARI} ($\uparrow$) & \textbf{NMI} ($\uparrow$) & \textbf{ASW} ($\uparrow$) & \textbf{cMCC-P} ($\downarrow$) & \textbf{cMCC-S} ($\downarrow$) \\ \midrule
\textbf{MO-CO-cVAE} & $\mathbf{0.34 \pm 0.05}$ & $\mathbf{0.40 \pm 0.03}$ & $\mathbf{0.10 \pm 0.01}$ & $\mathbf{0.28 \pm 0.04}$ & $\mathbf{0.28 \pm 0.04}$ \\ 
\textbf{SO-CO-cVAE} & $0.31 \pm 0.06$ & $0.38 \pm 0.03$ & $0.08 \pm 0.01$ & $\mathbf{0.28 \pm 0.02}$ & $\mathbf{0.28 \pm 0.02}$ \\
\textbf{MO-cVAE} & $0.32 \pm 0.09$ & $0.39 \pm 0.06$ & $0.07 \pm 0.05$ & $0.36 \pm 0.06$ & $0.36 \pm 0.06$ \\
\textbf{SO-cVAE} & $0.27 \pm 0.02$ & $0.31 \pm 0.04$ & $0.04 \pm 0.01$ & $0.40 \pm 0.01$ & $0.40 \pm 0.01$ \\
\textbf{ContrastiveVI} & $0.30 \pm 0.06$ & $0.38 \pm 0.05$ & $0.06 \pm 0.03$ & $0.34 \pm 0.04$ & $0.34 \pm 0.05$ \\
\textbf{cVAE} & $0.27 \pm 0.10$ &	$0.31 \pm 0.08$ & $0.05 \pm 0.04$ &	$0.36 \pm 0.02$ & $0.36 \pm 0.02$\\
\textbf{VAE} & $0.28 \pm 0.05$ & $0.34 \pm 0.05$ & $0.06 \pm 0.02$ & $0.76 \pm 0.10$ & $0.74 \pm 0.11$ \\ \bottomrule
\end{tabular}
}
\vspace{-3mm}
\end{table}

%% file: 8_discussion.tex
\section{Conclusion}
This study examines the identifiability properties of recently proposed Deep Generative Models (DGMs) for comparative analysis. Our analysis highlights the block-wise identifiability of many recent contrastive and multi-group DGMs, drawing connections between data from differing sources and the broader landscape of causal representation learning. A significant contribution is the extension of non-linear ICA results to count distributions, an area previously less explored. We further assess the challenges associated with estimating the number of latent variables and provide empirical evidence that regularization is beneficial under those specific circumstances. Building on our theoretical findings, we introduce a methodology grounded in multi-objective and constrained optimization principles. As the field continues to employ DGMs in diverse scientific applications, it is crucial to emphasize the dual objectives of accurate model fit and interpretability, ensuring that the generated models are both robust and scientifically valuable.

%% file: supplements.tex
\part*{Appendices}

\startcontents[sections]
\printcontents[sections]{l}{1}{\setcounter{tocdepth}{2}}

\newpage

\input{7_relatedwork}

\section{Multi-group DGMs}
\label{app:multi-group}
For the sake of completeness, we describe the framework of multi-group analysis~\citep{weinberger22disentangling}. We note that this framework has also been referred to as cross-population deep generative modeling in~\cite{davison2019cross}.

\subsection{Generative model}
When dealing with multiple data sources (focusing on two data sets for the sake of simplicity), one immediate question for data exploration is to enumerate patterns that are shared by all data sets versus the ones that appear only in one of the data sets but not the other. This is the objective of a Multi-Group analysis model~(Figure~\ref{fig:generative-models}, right). In this model, we have three blocks of latent variables. First, latent variable
\begin{align}
\z \sim \textrm{Normal}(0, I_p),
\end{align}
that encodes shared variation between the two data sets. Then, latent variable
\begin{align}
\ta \sim \textrm{Normal}(0, I_q),
\end{align}
encodes variation that is unique to the first data set. Similarly, 
\begin{align}
    \tb \sim \textrm{Normal}(0, I_q),
\end{align} 
encodes variation that is unique to the second data set. Observations $x$ are then drawn according to an exponential family distribution, with a mixing function $f$:
\begin{align}
\x \sim \textsc{ExpFam}\left(f(\z, \ta, \tb)\right).
\end{align}

\subsection{Data Set Definition}
Interestingly, we have no data available from the distribution $p_\theta(\x)$, because we observe data from either data set, where one of the variables is inactive ($\ta=\zero$ or $\tb=\zero$), as introduced in~\citet{weinberger22disentangling} as well as~\cite{davison2019cross}. To formalize this, we model it as a hard intervention~\citep{pearl2009causal}, akin to interventional causal representation learning~\citep{ahuja2023interventional}. For example, data set 1 is generated by sampling from the distribution $p_\theta\left(\x \mid \text{do}(\tb = \zero)\right)$, and we operate similarly for the data set 2, sampled from $p_\theta\left(\x \mid \text{do}(\ta = \zero)\right)$.  The reader will notice that the setting of contrastive analysis (Figure~\ref{fig:generative-models}, left) is a particular instance of this model, when the mixing function is constant with respect to one of the group-specific latent variables (e.g., $\ta$).

\subsection{Variational Inference}
Both of the data likelihoods $p_\theta\left(\x \mid \text{do}(\ta = \zero)\right)$ and $p_\theta\left(\x \mid \text{do}(\tb = \zero)\right)$ are intractable. Therefore, ~\citet{weinberger22disentangling} as well as~\cite{davison2019cross} both proceeded to posterior approximation with variational inference. 

For each sample in data set 1, the variational distribution is mean-field $q_\phi(\z \mid \x)q_\phi(\ta \mid \x)$. Then, the evidence lower bound (ELBO) is written as:
\[
    \mathcal{L}^1({\theta, \phi}) = \E_{q_\phi(\z \mid \x)q_\phi(\ta \mid \x)}\log \frac{p_\theta(\x, \ta, \zero)}{q_\phi(\z \mid \x)q_\phi(\ta \mid \x)}.
\]
For each sample in data set 2, the variational distribution is $q_\phi(\z \mid \x)q_\phi(\tb \mid \x)$. Then, the evidence lower bound is written as:
\[
    \mathcal{L}^2({\theta, \phi}) = \E_{q_\phi(\z \mid \x)q_\phi(\tb \mid \x)}\log \frac{p_\theta(\x, \zero, \tb)}{q_\phi(\z \mid \x)q_\phi(\tb \mid \x)}.
\]
Both frameworks propose to optimize the following composite ELBO: 
\[
    \mathcal{L}({\theta, \phi}) = \mathcal{L}^1({\theta, \phi}) + \mathcal{L}^2({\theta, \phi}).
\]
This composite ELBO may be used as an objective function for maximization, in conjunction with adequate regularization of the neural networks parameterizing the variational distribution as a function of the input data. In~\citet{weinberger2022moment}, the regularization ensures that the output of the neural network parameterizing the variational posterior for the latent variables $\ta$ and $\tb$ (i.e., their mean and diagonal variance vector) is close to zero for data points where the value of $\ta$ or $\tb$ should be zero. For example, for a point $x^1$ from data set 1, the regularization will penalize the sum of the square mean and the variance of $q_\phi(\tb \mid \x)$, which corresponds to the Wasserstein distance between $q_\phi(\tb \mid \x)$ and the Dirac distribution centered at $\zero$.  

\section{A Theory of Identifiability for Noiseless Comparative Analysis DGMs}
\subsection{Counterexample of Identifiability for Non-linear Contrastive Analysis}
\label{app:general-counterexample}
\paragraph{Data Generating Model} Let us assume we observe the target data according to the following contrastive deep generative model:
\begin{align}
    \z &\sim \textrm{Normal}(0, I_p)\\
    \s &\sim \textrm{Normal}(0, I_q)\\
    \x &\sim \textrm{Normal}(f(\z, \s), \sigma^2I_d).
\end{align}
We also observe a background data set from the distribution $p_\theta\left(\x \mid \text{do}(\s = \bm{0})\right)$. In this example, we consider $p=2$ and $q=1$ to demonstrate that there exist mixing functions $f$ that cannot be identified from data. Towards this end, we build a function $\tilde{f}$ such that the resulting data distribution is identical to that of $f$, but such that $\tilde{f}$ is not subspace disentangled with respect to $f$.

\paragraph{A Key Diffeomorphism}
The key idea for this counterexample consists in using a diffeomorphism of $\R^3$ that preserves volumes, as well as Euclidean distances to the origin. We consider the following diffeomorphism:
\begin{align}
\Phi: \left(\begin{array}{l}
z_1 \\ z_2 \\ s
\end{array}
\right) &\mapsto
\left(\begin{array}{c}
z_1 \cos s - z_2 \sin s \\ z_1 \sin s + z_2 \cos s \\ s
\end{array}
\right).
\end{align}
To verify that $\Phi$ is norm-preserving, we denote by $R_s$ the rotation operator in two dimensions and simply calculate
\begin{align}
    \forall z_1, z_2, s \in \R^3, \norm{\Phi(z_1, z_2, s)}^2_2 &= \norm{R_s((z_1, z_2)}^2_2 + s^2 = \norm{(z_1, z_2, s)}_2^2.
\end{align}
To verify that $\Phi$ is volume-preserving, we may calculate the Jacobian determinant:
\begin{equation}
\left|D_\Phi(z_1, z_2, s)\right| = \left|\begin{array}{ccc}
\cos s & -\sin s & 0 \\
\sin s & \cos s & 0 \\ 
0 & 0 & 1
\end{array}
\right| = 1.
\end{equation}
Next, we make use of the following lemma.

\begin{lem}
    Let $\Phi$ be a diffeomorphism of $\R^d$. Let us assume that $\Phi$ preserves the Euclidean norm, that is that for all $\bm{u} \in \R^d$, $\norm{\Phi(\bm{u})}_2 = \norm{\bm{u}}_2$. Let us also assume that $\Phi$ is volume-preserving, meaning that for all $\bm{u} \in \R^d$, $|D_\Phi(\bm{u})| \in \{-1, 1\}$. Then, $\Phi$ leaves the isotropic Gaussian distribution invariant, that is for $\x \sim \textrm{Normal}(0, I_d)$, $\x \equald \Phi(\x)$.
\end{lem}
\proof{
To show that $\x$ and $\tilde{\x} = \Phi(\x)$ are equal in distribution, we will show equality of the characteristic functions. 

Let $\tv \in \R^d$. The characteristic function $\phi_{\tilde{\x}}(\tv)$ of random variable $\tilde{\x}$ is defined as:
\begin{align}
    \phi_{\tilde{\x}}(\tv) = \E_{\tilde{\x}}e^{i\tv^\top\tilde{\x}}
    = \int e^{i\tv^\top\tilde{\x}}d\Prob_{\tilde{\x}}.
\end{align}
Using the change of variable formula, we have:
\begin{align}
    \phi_{\tilde{\x}}(\tv) &= \int_{\R^d} e^{i\tv^\top\tilde{\x}}|D_{\Phi^{-1}}(\tilde{\x})|p_{\x}(\Phi^{-1}(\tilde{\x}))d\tilde{\x}.
\end{align}
Now, because $\Phi$ is a volume-preserving diffeomorphism, we have $|D_{\Phi^{-1}}(\tilde{\x})| = 1$ for all $\tilde{\x} \in \R^d$. And, because $p_{\x}$ is the density of the isotropic Gaussian distribution, we have that $p_{\x}$ depends only on the distance to the origin (i.e., the Euclidean norm). However, $\Phi$ is norm-preserving so we have $p_{\x}(\Phi^{-1}(\tilde{\x})) = p_{\x}(\tilde{\x})$ for all $\tilde{\x} \in \R^d$. Therefore, we have:
\begin{align}
    \phi_{\tilde{\x}}(\tv) &= \int_{\R^d} e^{i\tv^\top\tilde{\x}}p_{\x}(\tilde{\x})d\tilde{\x} = \phi_{\x}(\tv),
\end{align}
which concludes the proof. 
\QEDB
}

\paragraph{Unidentifiability} Because any diffeomorphism that preserves norm and volume leaves the isotropic Gaussian distribution invariant, $\Phi(\z, \s) \equald (\z, \s)$. Furthermore, we notice that the restriction of $\Phi$ to the domain $\R^2 \times \{0\}$ satisfies the same properties, and therefore $\Phi(\z, 0) \equald (\z, 0)$. Now, for any non-trivial $f$,  we define $\tilde{f} = f \circ \Phi$ so that $v = \Phi$. We have constructed a case of two functions $f$ and $\tilde{f}$ where the data distributions are equal, but $\tilde{f}$ is not subspace disentangled with respect to $f$, because $v$ is not compatible with the Cartesian product $\R^2 \times \R$. Indeed, the first component of $v(\z, \s)$ depends non-trivially on $\s$.

\paragraph{Extension to the Multi-group setting} The reader will notice that this counterexample may be easily adapted to the multi-group setting, by rotating the block of $z$ by and angle of $\ta + \tb$. 

\subsection{Identifiability of ReLU Contrastive Analysis DGMs}
\label{app:relu}

We start by stating a general result from~\citet{kivva2022identifiability} that we will apply to the comparative analysis setting.

\begin{theorem}[Theorem D.4 from~\citet{kivva2022identifiability}]
    \label{thm:kivva}
    Let $f, g: \R^m \rightarrow \R^n$ be continuous piecewise affine functions such that $f$ and $g$ are both injective for almost every point in their respective images $f(\R^m)$ and $g(\R^m)$. Let $Z \sim \sum_{i=1}^J \lambda_i\textrm{Normal}(\mu_i, \Sigma_i)$ and $Z' \sim \sum_{j=1}^{J'} \lambda_j\textrm{Normal}(\mu'_j, \Sigma'_j)$ be a pair of variables with GMM distribution (in reduced form). Suppose that $f(Z)$ and $g(Z')$ are equally distributed. Let $\mathcal{D} \subseteq \R^n$ be a connected open set such that $f$ and $g$ are injective onto $\mathcal{D}$. Then, there exists an affine transformation $h: \R^m \rightarrow \R^m$ such that $h(Z) \equald Z'$ and $g(z) = \left(f \circ h^{-1}\right)(z)$ for every $z \in g^{-1}(\mathcal{D})$.
\end{theorem}

We now prove our main result.

\theorelu*
\proof{
Let $f$ and $\tilde{f}$ be two continuous injective piecewise linear functions such that $f(\z, \s) \equald \tilde{f}(\z, \s)$ and $f(\z, \zero) \equald \tilde{f}(\z, \zero)$. The key idea of the proof is to first study the implications of each of the two hypotheses on equality in distribution independently. By applying the result of \cite{kivva2022identifiability}, Theorem~\ref{thm:kivva}, we reduce the complexity of the problem to a linear equivalence class of functions. Then, simple linear algebra allows us to conclude. 

We start by applying the result of~\cite{kivva2022identifiability} to the target data set, as it is the most straightforward. $f$ and $\tilde{f}$ are two continuous injective piecewise affine functions. Because for the Gaussian vector $(\z, \s)$, we have $f(\z, \s) \equald \tilde{f}(\z, \s)$, we may apply Theorem~\ref{thm:kivva} and obtain that there exists an affine transformation $h_1: \R^p \times \R^q \rightarrow \R^p \times \R^q$ such that $h_1(\z, \s) \equald (\z, \s)$, and
\begin{align}
    \forall (\uvar, \vv) \in \R^p \times \R^q, \tilde{f}(\uvar, \vv) = \left(f \circ h_1^{-1}\right)(\uvar, \vv).
\end{align}

Now, let us consider the background data set. Let $\mathcal{D}^b =\R^p \times \{0\}$ denote the domain of the latent variables that generate the background data set, and let $\restr{f}{\mathcal{D}^b}$ be the the restriction of the piecewise affine function $f$ to the domain $\mathcal{D}^b$. Because $f$ is piecewise affine, $\restr{f}{\mathcal{D}^b}$ is also piecewise affine. Because the restriction of an injective function is injective, $\restr{f}{\mathcal{D}^b}$ is injective. Given that $f(\z, \zero) \equald \tilde{f}(\z, \zero)$, we may apply Theorem~\ref{thm:kivva} and obtain that there exists an affine transformation $h_0: \R^p \rightarrow \R^p$ such that $h_0(\z) \equald \z$, and
\begin{align}
    \forall \uvar \in \R^p, \restr{\tilde{f}}{\mathcal{D}^b}(\uvar, \zero) = \left(\restr{f}{\mathcal{D}^b} \circ h_0^{-1}\right)(\uvar, \zero).
\end{align}
Finally, we appeal to a short linear algebraic argument to conclude. Because $h_0$ and $h_1$ are affine maps preserving the isotropic Gaussian distribution, they must be linear (i.e., the offset term is zero) in order to preserve the mean. We may write the functions as
\begin{align}
    h_0(\uvar) &= R_0\uvar\\
    h_1(\uvar, \vv) &= \left(R_{11}\uvar + R_{12}\vv, R_{21}\uvar + R_{22}\vv\right).
\end{align}
Now, because of the injectivity of the functions $f$ and $\tilde{f}$, we know that $h_0$ and $h_1$ are equal for all values of $\uvar$ whenever $\vv=\zero$. Therefore, we have that $R_{11} = R_0$, and $R_{21} = 0$. We may rewrite our functions as
\begin{align}
    h_0(\uvar) &= R_0\uvar\\
    h_1(\uvar, \vv) &= \left(R_0\uvar + R_{12}\vv, R_{22}\vv\right).
\end{align}
Now, in order to preserve the covariance of the Gaussian vector, we have
\begin{equation}
    \begin{pmatrix}
        I_p & 0\\0 & I_q
    \end{pmatrix}
    =
    \begin{pmatrix}
        R_0 & R_{12}\\
        0 & R_{22}
    \end{pmatrix}
    \begin{pmatrix}
        R_0^\top & 0\\
        R_{12}^\top & R_{22}^\top
    \end{pmatrix}
    =
    \begin{pmatrix}
        R_0 R_0^\top + R_{12} R_{12}^\top & R_{12} R_{22}^\top\\
        R_{22} R_{12}^\top & R_{22} R_{22}^\top
    \end{pmatrix}.
\end{equation}
Therefore, we have that $R_{22}R_{22}^\top = I$, and $R_{22}$ is an orthogonal matrix. But additionally, $R_{12}R_{22}^\top = 0$, and because $R_{22}$ is invertible, it implies that $R_{12} = 0$.
Therefore $R_{12}$ is the null function and we have proved that $v = f^{-1} \circ \tilde{f}$ is compatible with respect to the Euclidean decomposition $\R^p \times \R^q$.
\QEDB}

\subsection{Identifiability of ReLU Multi-Group Analysis DGMs}
\label{app:multi-group-proofs}

We start this section by introducing the notation and the definition of subspace disentanglement in this specific setting. Let us also introduce the domains for the latent variables $\mathcal{D}^1 = \mathcal{Z} \times \mathcal{T}^1 \times \{0\}$ and $\mathcal{D}^2 =\mathcal{Z} \times \{0\} \times  \mathcal{T}^2$. The subspace identifiability corresponds to:
\begin{definition}[Subspace Disentanglement]
A learned mixing function $\tilde{f}$ is said to be subspace-disentangled with respect to the ground truth mixing function $f$ if $f(\mathcal{D}^1) = \tilde{f}(\mathcal{D}^1)$ and $f(\mathcal{D}^2) = \tilde{f}(\mathcal{D}^2)$ and the mapping $v = f^{-1} \circ \tilde{f}$ is a diffeomorphism such that both $\restr{v}{\mathcal{D}^1}$ and $\restr{v}{\mathcal{D}^2}$ are compatible with respect to the subspaces $\mathcal{D}^1$ and $\mathcal{D}^2$.
\end{definition}
When this property is not verified, this would imply that the learned mixing function $\tilde{f}$ does not decompose the signal in the feature space $\mathcal{X}$ into the same shared $\mathcal{Z}$ and private spaces $\mathcal{T}^1$, and $\mathcal{T}^2$. Now we proceed to a statement of the generalization of Theorem~\ref{theo:relu} to the multi-group setting.
\begin{theo}
\label{theo:relu-group}
Let the ground truth mixing function $f$ and the learned mixing function $\tilde{f}$ both be continuous and injective piecewise affine mixing functions such that $f(\z, \ta, \zero) \equald \tilde{f}(\z, \ta, \zero)$ and $f(\z, \zero, \tb) \equald \tilde{f}(\z, \zero, \tb)$. Then, $\tilde{f}$ is subspace disentangled with respect to $f$ and the noiseless version of the multi-group analysis model is subspace identifiable.
\end{theo}
\proof{
Let $f$ and $\tilde{f}$ be two continous injective piecewise linear functions such that
\begin{equation}
    f(\z, \ta, \zero) \equald \tilde{f}(\z, \ta, \zero) \quad
    \text{and} \quad
    f(\z, \zero, \tb) \equald \tilde{f}(\z, \zero, \tb).
\end{equation}
We proceed very similar to the contrastive analysis case. 

Let $\mathcal{D}^1 =\R^p \times \R^q \times \{0\} $ denote the domain for the latent variables that generate data set 1. Let $\restr{f}{\mathcal{D}^1}$ be the the restriction of the piecewise affine function $f$ to the domain $\mathcal{D}^1$. Because $f$ is piecewise affine, $\restr{f}{\mathcal{D}^1}$ is also piecewise affine. Because the restriction of an injective function is injective, $\restr{f}{\mathcal{D}^1}$ is injective. Given that $f(\z, \ta, \zero) \equald \tilde{f}(\z, \ta, \zero)$, we may apply Theorem~\ref{thm:kivva} and obtain that there exists an affine transformation $h_1: \mathcal{D}^1 \rightarrow \mathcal{D}^1$ such that $h_1(\z, \ta, \zero) \equald (\z, \ta, \zero)$, and:
\begin{align}
    \forall \uvar, \vv^1 \in \R^p \times \R^q, \restr{\tilde{f}}{\mathcal{D}^1}(\uvar, \vv^1, \zero) = \left(\restr{f}{\mathcal{D}^B} \circ h_1^{-1}\right)(\uvar, \vv^1, \zero).
\end{align}
Applying the same argument to data set 2, we conclude that that there exists an affine transformation $h_2: \mathcal{D}^2 \rightarrow \mathcal{D}^2$ such that $h_2(\z, \zero, \tb) \equald (\z, \zero, \tb)$, and:
\begin{align}
    \forall \uvar, \vv^2 \in \R^p \times \R^q, \restr{\tilde{f}}{\mathcal{D}^1}(\uvar, \zero, \vv^2) = \left(\restr{f}{\mathcal{D}^B} \circ h_2^{-1}\right)(\uvar, \zero, \vv^2).
\end{align}
Finally, we make appeal to a short linear algebraic argument to conclude. Because $h_1$ and $h_2$ are affine maps preserving the isotropic Gaussian distribution, they must be linear in order to preserve the mean. We may rewrite the definitions of the functions as:
\begin{align}
    h_1(\z, \vv^1) &= \left(U_{11}\z + U_{12}\vv^1, U_{21}\z + U_{22}\vv^1\right)\\
    h_2(\z, \vv^2) &= \left(V_{11}\z + V_{12}\vv^2, V_{21}\z + V_{22}\vv^2\right).
\end{align}
Due to the identifiability of the functions, we know that $h_1$ and $h_2$ overlap for all values of $\z$ when $\vv^1=\vv^2=\zero$. Therefore, we have that $U_{11} = V_{11} = R_1$, and $U_{21} = V_{21} = 0$. Therefore, we have 
\begin{align}
    h_1(\z, \vv^1) &= \left(R_1\z + U_{12}\vv^1, U_{22}\vv^1\right)\\
    h_2(\z, \vv^2) &= \left(R_{1}\z + V_{12}\vv^2, V_{22}\vv^2\right).
\end{align}
Now, in order to preserve the covariance of the Gaussian vector $h_1(\z, \tv^1)$, we have
\begin{equation}
    \begin{pmatrix}
        I_p & 0\\0 & I_q
    \end{pmatrix}
    =
    \begin{pmatrix}
        R_1 & U_{12}\\
        0 & U_{22}
    \end{pmatrix}
    \begin{pmatrix}
        R_1^\top & 0\\
        U_{12}^\top & U_{22}^\top
    \end{pmatrix}
    =
    \begin{pmatrix}
        R_1 R_1^\top + U_{12} U_{12}^\top & U_{12} U_{22}^\top\\
        U_{22} U_{12}^\top & U_{22} U_{22}^\top
    \end{pmatrix}.
\end{equation}
Observing that $U_{22}$ is an isometry, as in the contrastive case, we obtain that $U_{12} = 0$. Proceeding similarly for $h_2(\x, \tv^1)$, we have that $V_{12} = 0$ as well. 
Therefore, we have proved that the restrictions of $v = f^{-1} \circ \tilde{f}$ to $\mathcal{D}^1$ and $\mathcal{D}^2$ are compatible with respected to the Euclidean decomposition $\R^p \times \R^q \times \R^q$.
\QEDB}

\section{Identifiability Theory for Counting Observation Noise}

\subsection{Preliminary Results}
The central argument for our proofs will rely on a generalization of the characteristic function that we refer to as the Laplace transform:
\begin{definition}
    Let $\y$ be a random vector valued in $\R^d$. The Laplace transform of the random vector $\y$ is defined as:
    \begin{align}
        \mathbb{C}^d &\rightarrow \mathbb{C}\notag\\
        \xi_{\y}: \tv &\mapsto \E\left[e^{\tv^\top \y}\right].
    \end{align}
\end{definition}
The restriction of $\xi_{\y}$ to the product of the imaginary lines is the characteristic function $\phi_{\y}$. Similarly, the restriction of $\xi_{\y}$ to the product of the real lines is the moment generating function.

When dealing with count distributions, it is especially interesting to consider specific results about positive random variables. In this case, the Laplace transform is defined, and even holomorphic, over the product of half-spaces:
\begin{lemma}
\label{lemma:holomorphy}
    Let $\y$ be a random vector valued in $\R_+^d$. Then, the function $\xi_{\y}$ is a holomorphic function of several variables on $\mathcal{H}_{-}^d$, where $\mathcal{H}_{-} = \left\{z \in \mathbb{C} \mid \Re(z) < 0\right\}$.
\end{lemma}
\proof{We first show that the function is well defined on its domain $\mathcal{H}_{-}^d$. Let $\bm{t} = (u_j + iv_j)^d_{j=1}$ with $u_j <0$ for all $j \in [d]$. In this case, the integral is absolutely convergent:
\begin{align}
\label{eq:holomorph}
     \E\left|e^{\tv^\top \y}\right| &= \E\left|e^{\uvar^\top \y + i\vv^\top \y}\right| = \E e^{\uvar^\top \y} \leq 1
\end{align}
Because $\y$ is positive almost surely, and $\uvar$ is a vector with negative values. To show that $\xi_{\y}$ is holomorphic on $\mathcal{H}_{-}^d$, by Osgood's lemma, it is sufficient to prove that $\xi_{\y}$ is continuous on $\mathcal{H}_{-}^d$, and holomorphic in each of its variables on $\mathcal{H}_{-}$. To show that $\xi_{\y}$ is continuous, it is enough to notice that the integrand defining $\xi_{\y}$ is uniformly bounded by an integrable function (Equation~\ref{eq:holomorph}), and that the function $\tv \mapsto e^{\tv^\top\y}$ is continuous, so continuity follows from the dominated convergence theorem. To show that $\xi_{\y}$ is holomorphic on each of its variable, it is enough to show that $t_1 \mapsto \xi_{\y}(t_1, \ldots, t_d)$ is holomorphic on $\mathcal{H}_{-}$. Here again, we may apply the dominated convergence theorem. This is justified because (i) the exponential function is holomorphic,  (ii) for all $t_1 \in \mathcal{H}^{-}$ the integral exists and (iii) the integrand is uniformly bounded above by an absolutely integrable function. 
\QEDB}

Next, we derive the Laplace transform for some specific compound variables. 

\begin{lemma}[Poisson noise]
\label{lemma:poisson}
    Let $\y$ be a random vector valued in $\R_+^d$, and let us assume that we observe count data $\x$ generated as $x_j \sim \textrm{Poisson}(y_j)$ for $j \in [d]$. Then, for all $\tv \in \C^d$ such that the Laplace transform of the random vector $\x$ is defined, it can be derived as:
    \begin{align}
        \xi_{\x}(\tv) = \xi_{\y}\left(e^{\tv} - \bm{1}\right),
    \end{align}
    where the component-wise operations are used to assemble the vector $e^{\tv} - \bm{1} = \left(e^{t_j} - 1\right)_{j=1}^d$.
\end{lemma}
\proof{
This derivation simply uses the law of total expectations,
\begin{align}
    \xi_{\x}(\tv) &= \E_{\x}[e^{\tv^\top \x}]= \E_{\y}\left[\E_{\x}[e^{\tv^\top \x}\mid \y]\right],
    \end{align}
the fact that components of $\x$ are independent conditionally on $\y$,
\begin{align}
    \xi_{\x}(\tv) &= \E_{\y}\left[\E_{\x}\left[\prod_{j=1}^d e^{t_j x_j}\mid \y\right]\right] = \E_{\y}\left[\prod_{j=1}^d\E_{x_j}\left[e^{t_j x_j}\mid \y\right]\right] ,
    \end{align}
and the definition of the Laplace transform for the Poisson distribution
\begin{align}
    \xi_{\x}(\tv) = \E_{\y}\left[\prod_{j=1}^d e^{y_j(e^{t_j} -1)}\right] = \xi_{\y}\left(e^{\tv} - \bm{1}\right),
\end{align}
where component-wise operations are used to assemble the vector $e^{\tv} - \bm{1} = \left(e^{t_j} - 1\right)_{j=1}^d$.
\QEDB}

\begin{lemma}[Negative binomial noise]
\label{lemma:nb}
Let $\y$ be a random vector valued in $\R_+^d$, and let us assume that we observe count data $\x$ generated as $x_j \sim \textrm{NegativeBinomial}(y_j, \theta)$ for $j \in [d]$, where $\theta$ designates the shape parameter. Then, for all $\tv \in \C^d$ such that the Laplace transform of the random vector $\x$ is defined, it can be derived as:
\begin{align}
\xi_{\x}(\tv) &= \xi_{\y}\left(-\log (1 - (e^{\tv} -1)\theta)\right),
\end{align}
where $\log$ is the principal branch of the complex logarithm, and component-wise operations are used to assemble the vector $\log (1 - (e^{\tv} -1)\theta) = \left(\log (1 - (e^{t_j} -1)\theta)\right)_{j=1}^d$.
\end{lemma}
\proof{
In this case, we use the definition of the negative binomial distribution as a Gamma-Poisson compound distribution:
\begin{align}
    x_j &\sim \textrm{NegativeBinomial}(y_i, \theta) \Leftrightarrow  u_i \sim \textrm{Gamma}(y_i, \theta) , x_i \sim \textrm{Poisson}(u_i).
\end{align}
Following the Poisson case, we get
\begin{align}
    \xi_{\x}(\tv) &= \E_{\y}\left[\E_{\bm{u}}\left[\E_{\x}\left[e^{\tv^\top \x}\mid \bm{u}\right]\mid \y \right]\right]\\
     &= \E_{\y}\left[\E_{\bm{u}}\left[\E_{\x}\left[\prod_{j=1}^d e^{t_j x_j}\mid \bm{u}\right]\mid \y \right]\right]\\
     &= \E_{\y}\left[\prod_{j=1}^d\E_{\bm{u}}\left[\E_{\x}\left[ e^{t_j x_j}\mid \uvar \right] \mid \y\right]\right].
\end{align}
Then, using the definition of the Laplace transform for the Poisson distribution and the Gamma distribution, we get
\begin{align}
     \xi_{\x}(\tv)  &= \E_{\y}\left[\prod_{j=1}^d \E_{\uvar}e^{u_j(e^{t_j} -1)}\right]\\
     &= \E_{\y}\left[\prod_{j=1}^d \left[1 - (e^{t_j} -1)\theta\right] ^ {-y_j}\right],
\end{align}
where the power with a complex base $a^p$ is defined as $e^{p \log a}$, where $\log$ is the principal branch of the complex logarithm. Reassambling the product into a sum of terms inside the exponential distribution, we get
\begin{align}
\xi_{\x}(\tv) &= \xi_{\y}\left(-\log (1 - (e^{\tv} -1)\theta)\right),
\end{align}
where the vector $\log (1 - (e^{\tv} -1)\theta) = \left(\log (1 - (e^{t_j} -1)\theta)\right)_{j=1}^d$ is put together component-wise.
\QEDB}

\subsection{Identifiability of mixture through observational count noise}

\label{app:poisson-nb}
\begin{restatable}[Identifiability of mixture through observational count noise]{prop}{propmoment}
    \label{prop:moment}
    Let $\uvar$ be a random variable taking values in $\R^p$. Let $f$ and $\tilde{f}$ be two functions valued in $\R_+^d$, and let $\bm{y} = f(\uvar)$ and $\tilde{\bm{y}} = \tilde{f}(\uvar)$. If the random variables $\x \sim p_x(\bm{y})$ and $\tilde{\x} \sim p_x(\tilde{\bm{y}})$ are equal in distributions, and $p_x$ is Poisson, or negative binomial with fixed shape, then  it follows that $\bm{y} \equald \tilde{\bm{y}}$. 
\end{restatable}
\proof{
Because both random vectors $\y$ and $\tilde{\y}$ are valued in $\R_+^d$, Lemma~\ref{lemma:holomorphy} dictates that $\xi_{\y}$ and $\xi_{\tilde{\y}}$ are holomorphic on $\mathcal{H}_{-}^d$. Proceeding similarly for $\x$ and $\tilde{\x}$, $\xi_{\x}$ and $\xi_{\tilde{\x}}$ are holomorphic, and therefore well-defined on $\mathcal{H}_{-}^d$. 

Under the assumption that $\x \equald \tilde{\x}$, we know that the characteristic functions of $\x$ and $\tilde{\x}$ must agree. 
Therefore, we have:
\begin{align}
\label{eq:char}
    \forall \tv \in (i\R)^d, \xi_{\x}(\tv) = \xi_{\tilde{\x}}(\tv),
\end{align}
We now split the cases of Poisson and negative binomial distribution.

\textbf{Poisson case}~
Using Lemma~\ref{lemma:poisson}, Equation~\ref{eq:char} is equivalent to:
\begin{align}
    \forall \tv \in (i\R)^d, \xi_{\y}\left(e^{\tv}- \bm{1}\right) = \xi_{\tilde{\y}}\left(e^{\tv}- \bm{1}\right),
\end{align}
Therefore, the holomorphic functions $\xi_{\y}$ and $\xi_{\tilde{\y}}$ have the same value on a sequence of points $\tv_n = t_n \cdot \bm{1}_d$ where  $t_n = e^{i(\pi + \frac{1}{n})} -1$, that converges to $-2 \cdot \bm{1}_d \in \mathcal{H}^d_{-}$. 

\paragraph{Negative binomial case}
Using Lemma~\ref{lemma:nb}, Equation~\ref{eq:char} is equivalent to:
\begin{align}
    \forall \tv \in (i\R)^d, \xi_{\y}\left(-\log (1 - (e^{\tv} -1)\theta)\right) = \xi_{\tilde{\y}}\left(-\log (1 - (e^{\tv} -1)\theta)\right),
\end{align}
Therefore, the holomorphic functions $\xi_{\y}$ and $\xi_{\tilde{\y}}$ have the same value on a sequence of points $\tv_n = t_n \cdot \bm{1}_d$ where  $t_n = -\log\left(1 - \theta(e^{i(\pi + \frac{1}{n})} -1)\right)$, that converges to $-\log(1 + 2\theta) \cdot \bm{1}_d \in \mathcal{H}^d_{-}$. 

\textbf{End of proof}~
In both cases, by analytic continuation, we have:
\begin{align}
    \forall \tv \in \mathcal{H}^d_{-}, \xi_{\y}(\tv) = \xi_{\tilde{\y}}(\tv).
\end{align}
Now, let us notice that $\xi_{\y}(\tv)$ and $\xi_{\tilde{\y}}(\tv)$ are holomorphic, and therefore continuous on $\mathcal{H}^d_{-}$. 
For each $\bm{w} \in \R^d$, we denote as $\tv_n^{\bm{w}} = -\frac{1}{n} \cdot \bm{1}_d + i\bm{w}$. We know that both functions $\xi_{\y}$ and $\xi_{\tilde{\y}}$ admit for limit the value of their respective characteristic function evaluated at $\bm{w}$ when $n \rightarrow \infty$, and then by continuous extension of the function, we must have equality of the limits, and therefore of the characteristic functions. Therefore, this is enough to guarantee $\y \equald \tilde{\y}$, since characteristic functions uniquely characterize a probability distribution.
\QEDB}

\label{app:poissonaffine}

We now proceed to the proof of the theorem.

\theocounting*
\proof{The implication $\tilde{\x}\equald \x \implies \tilde{f}(\uvar) \equald f(\uvar)$ is the result of Proposition~\ref{prop:moment}. 

To prove the second implication, we need to notice that because $\sigma:\R \rightarrow \R$ is a bicontinuous bijection, it must be a monotonic function. Without loss of generality, we may assume it is strictly increasing. Therefore, its inverse function $\sigma^{-1}$ is also strictly increasing, and by monotonicity, for every line segment $[a, b]$ of $\R$, we have:
\begin{align}
    \forall c \in \R, c \in [a, b] \Leftrightarrow \sigma^{-1}(c) \in [\sigma^{-1}(a), \sigma^{-1}(b)].
\end{align}
Let us consider now a hyperbox $H = \prod_{i=1}^d [a_i, b_i]$, and $H' = \prod_{i=1}^d [\sigma^{-1}(a_i), \sigma^{-1}(b_i)]$ its image component-wise by $\sigma$. We have again, for every point $\bm{c} \in \R^d$:
\begin{align}
    \bm{c} \in H \Leftrightarrow \Psi(\bm{c}) \in H',
\end{align}
with $\Psi$ is defined as
\begin{align}
    \Psi: \y \mapsto \left(\sigma^{-1}(y_1), \ldots, \sigma^{-1}(y_d)\right).
\end{align}
To conclude the proof, it is enough to notice that two random variables $X$ and $Y$ of $\R^d$ are equal in distributions if they have the same generalized cumulative distribution function, that is if for every hyperbox $H$, we have $p(X \in H) = p(Y \in H)$.

Indeed, if we assume that $\tilde{f}(\uvar) \equald f(\uvar)$, we have that for every hyperbox $H$, $p(\tilde{f}(\uvar) \in H) = p(f(\uvar) \in H)$. However, we have that $\tilde{f}(\uvar) \in H$ if and only if $\tilde{g}(\uvar) \in H'$, and similarly, $f(\uvar) \in H$ if and only if $g(\uvar) \in H'$. Therefore, we have that for every image hyperbox $H'$, $p(\tilde{f}(\uvar) \in H') = p(f(\uvar) \in H')$. Because $\sigma$ is a bicontinuous bijection, the set of images hyperboxes covers the set of all hyperboxes (because of monotonicity), and therefore, we have that $\tilde{g}(\uvar) \equald g(\uvar)$.
\QEDB}

\subsection{Discussion of the Bernoulli Noise Setting}
\label{app:ber}

The intuition is that in the one-dimensional setting, a mixture of Bernoulli distributions is still a distribution on $\{0,1\}$ and therefore entirely determined by its first moment. To show this, we proceed with calculations similar to the ones conducted for the Poisson and negative binomial distribution, but in the case of Bernoulli noise. Let $\y$ be a random vector valued in $[0, 1]^d$, and let us assume that we observe binary data $\x$ generated as $x_j \sim \textrm{Bernoulli}(y_j)$ for $j \in [d]$. For $\tv \in \C^d$, the Laplace transform of the random vector $\x$ can be derived as:
\begin{align}
    \xi_{\x}(\tv) &= \E_{\y}\left[\prod_{j=1}^d\E_{x_j}\left[ e^{t_j x_j}\mid \y\right]\right].
\end{align}
Let us notice that because $\y$ is bounded, $\xi_{\x}$ is defined on all of $\C^d$. Then, considering the definition of the Laplace transform of a Bernoulli distribution, we obtain
\begin{align}
     \xi_{\x}(\tv) &= \E_{\y}\left[\prod_{j=1}^d\left( (1 - y_j) + y_j e^{t_j}\right)\right],
\end{align}
and we may expand the product, by identifying the subset $S \subset \{1, \ldots, d\}$ with a binary vector of size $d$
    \begin{align}
     \xi_{\x}(\tv) &= \E_{\y}\left[\sum_{S \subset \{1, \ldots, d\}} \prod_{j=1}^d (1 - y_j)^{(1- S_j)}y_j^{S_j} e^{t_jS_j}\right]\\
     \xi_{\x}(\tv) &= \sum_{S \subset \{1, \ldots, d\}} \E_{\y}\left[\prod_{j=1}^d (1 - y_j)^{(1- S_j)}y_j^{S_j}\right] e^{\tv^\top S}.
\end{align}
It is important to notice here that each subset $S$ in the sum above induces a unique monomial term. Give this observation, if we assume $\xi_{\x} = \xi_{\tilde{\x}}$ for another data-generating process, the equality of those two functions implies the equality of two finite complex Fourier series, and therefore their coefficients must be equal (in fact, it is an equivalence):
\begin{align}
&\forall \tv \in \C^d, \xi_{\x}(\tv) = \xi_{\tilde{\x}}(\tv).\\
&\Leftrightarrow \forall S \in \{0, 1\}^d, \E_{\y}\left[\prod_{j=1}^d (1 - y_j)^{(1- S_j)}y_j^{S_j}\right] =  \E_{\tilde{\y}}\left[\prod_{j=1}^d (1 - y_j)^{(1- S_j)}y_j^{S_j}\right].
\label{eq:equivalence_bern}
\end{align}
Interestingly, the polynomial appearing in the last equation only has term with partial degree of at most 1 (but of total degree $d$). Therefore, we hypothesize that it must not be true that this fully characterizes the distribution of $Y$, because in general, there exist several distinct functions $f$ that could have the same moment of order $1$.

We therefore focus on building a counter-example with $p=d=2$. In this case, let us assume $\z \sim \textrm{Normal}(0, I_2)$ and that $\y = f(\z)$, and $\tilde{\y} = \tilde{f}(\z)$, and $\x$ is generated as $x_j \sim \textrm{Bernoulli}(y_j)$ for $j \in [d]$. $\tilde{\x}$ is generated similarly, from $\tilde{\y}$. Let us also assume $\tilde{\x} \equald \x$. Based on Equation~\ref{eq:equivalence_bern}, we have that
\begin{equation}
\left\{\begin{array}{lcl}
\E_{\y}\left[(1-y_1)(1-y_2)\right] &=& \E_{\tilde{\y}}[(1-\tilde{y}_1)(1-\tilde{y}_2)]\\
\E_{\y}[y_1(1-y_2)] &=& \E_{\tilde{\y}}[\tilde{y}_1(1-\tilde{y}_2)]\\
\E_{\y}[(1-y_1)y_2] &=& \E_{\tilde{\y}}[(1-\tilde{y}_1)\tilde{y}_2]\\
\E_{\y}[y_1y_2] &=& \E_{\tilde{\y}}[\tilde{y}_1\tilde{y}_2]
\end{array}
\right.
\end{equation}
These equations are equivalent to the constraints:
\begin{equation}
\label{eq:bern_constraints}
\left\{\begin{array}{lcl}
\E_{\y}\y &=& \E_{\tilde{\y}}\tilde{\y}\\
\E_{\y}y_1y_2 &=& \E_{\tilde{\y}}\tilde{y}_1\tilde{y}_2
\end{array}
\right.
\end{equation}
It is possible to find examples of functions that will yield distributions $\y$ and $\tilde{\y}$ that satisfy the constraints in Equation~\ref{eq:bern_constraints}, but have different distributions. For example, let us consider the functions $f_\lambda$ for $\lambda >0$ of the form:
\begin{align}
    f_\lambda: (z_1, z_2) \mapsto \left(F_\textrm{Beta}(F^{-1}_\textrm{Normal}(z_1); \lambda, \lambda), F_\textrm{Beta}(F^{-1}_\textrm{Normal}(z_2); \lambda, \lambda) \right),
\end{align}
where $F_\textrm{Normal}$ denotes the cumulative distribution function (CDF) of the isotropic Gaussian distribution, and $F_\textrm{Beta}(., \lambda, \lambda)$ denotes the CDF of the Beta  distribution with parameters $(\lambda, \lambda)$. For $\y = f_1(\z)$ and $\tilde{\y} = f_2(\z)$, we do not have equality in distribution for $\y$ and $\tilde{\y}$, but we do have that $\x \equald \tilde{\x}$.

To show that this pathological case can also happen within the framework we consider in this paper, let us consider the case $f = \sigma \circ g$ where $g$ is a piecewise affine function. Furthermore, we consider the case where $f$ is separable:
\begin{align}
    f(z_1, z_2) = \left(\sigma (g_{a, b}(z_1)), \sigma(g_{a, b}(z_2))\right),
\end{align}
where $\sigma$ denotes the sigmoid function, and
\begin{align}
    g_{a, b}(w) = aw\delta\{w \leq 0\} + bw\delta\{w > 0\}.
    \end{align}
Because $y_1$ and $y_2$ are independent in this case, equality in the mean of each component of $\y$ is enough to guarantee equality in distribution of $\y$. Let us examine the function:
\begin{align}
    \Phi: (a, b) \mapsto \E\left[\sigma(g_{a, b}(u))\right] = \int_{-\infty}^0 \sigma(aw)p(w)dw + \int^{+\infty}_0 \sigma(bw)p(w)dw,
\end{align}
where $u$ is a random variable distributed according to a standard Gaussian distribution, and $p(w)$ denotes its density evaluated at $w$. By continuity under the integral sign, $\Phi$ is a continuous function on its domain. 

We note that $\Phi$ cannot be injective, as it otherwise could be used to define an homeomorphism from $\R^2$ to $\R$, which is impossible by the invariance of domain theorem. Therefore, we must have two distinct parameters $(a, b)$ and $(a', b')$ such that the induced functions $f$ and $f'$ generate the same distribution. Because each set of parameters designates a unique function (if two $(a, b) \neq (a', b')$, then $f_{a, b} \neq f_{a', b'}$), this is a counterexample.

\section{Identifiability under Misspecification of Contrastive DGMs}

\subsection{Block-wise Identifiability of Misspecified Linear Model}
\label{app:identif_misspec_linear_model}
\propidentlinear*
\proof{
Let us assume that the data is generated according to the contrastive analysis model, and where $f: (\z, \s) \mapsto \x = U\z + V\s $ is a linear mixing function. Let $\tilde{f}: (\tilde{\z}, \tilde{\s}) \mapsto \tilde{\x} = \tilde{U}\tilde{\z} + \tilde{V}{\tilde{\s}} $ denote the learned mixing function. 

We first note that $f$ is a linear function, and $\z, \s$ is a Gaussian vector. Similarly, $\tilde{f}$ is a linear function, and $\tilde{\z}, \tilde{\s}$ are Gaussian vectors. Therefore, we have that $f(\z, \s)$ and $\tilde{f}(\tilde{\z}, \tilde{\s})$ are Gaussian vectors. Two Gaussian vectors are equal in distributions if they have the same mean and covariance matrix. Because both are centered (with mean zero), we therefore rely on the equality of their covariance matrices. We may proceed similarly for the random vectors  $f(\z, \zero)$ and $\tilde{f}(\tilde{\z}, \zero)$

Because all of the random vectors $\z, \s, \tilde{\z}, \tilde{\s}$ follow an isotropic Gaussian distribution, the equality of the covariance matrices entails that
\begin{equation}
\left\{\begin{array}{lcl}
U U^\top &=& \tilde{U} \tilde{U}^\top \\
U U^\top + V V^\top &=& \tilde{U} \tilde{U}^\top + \tilde{V} \tilde{V}^\top
\end{array}
\right.,
\end{equation}
equivalent to 
\begin{equation}
\left\{\begin{array}{lcl}
U U^\top &=& \tilde{U} \tilde{U}^\top \\
V V^\top &=& \tilde{V} \tilde{V}^\top
\end{array}
\right..
\end{equation}
Based on identifiability of the factor analysis model, \citep{shapiro1985identifiability}, we know that there exists two matrices $O_1$ and $O_2$ with orthogonal rows such that 
\begin{equation}
\left\{\begin{array}{lcl}
\tilde{U} &=& UO_1 \\
\tilde{V} &=& VO_2
\end{array}
\right.
\end{equation}
and that we have $\text{rank}(U) = \text{rank}(\tilde{U}) = p$ and $\text{rank}(V) = \text{rank}(\tilde{V}) = q$. 

We now seek to compute $v^{\dagger}= f^{-1} \circ \tilde{f}$. Let us notice that $f$ is an injective linear function, and therefore $f^{-1}(\x) = (W_{\z}\x, W_{\s}\x)$, where $W = [W_{\z}^\top, W_{\s}^\top]^\top$ is the pseudo-inverse of $[U, V]$. Substituting this into the expression for $v$, we have
\begin{align}
    v(\tilde{\z}, \tilde{\s}) = \left(W_{\z}UO_1\tilde{\z} + W_{\z}VO_2\tilde{\s}, W_{\s}UO_1\tilde{\z} + W_{\s}VO_2\tilde{\s}\right).
\end{align}
By definition of the pseudo-inverse, we have that $W_{\z}U = I$, $W_{\s}V = I$, $W_{\z}V = 0$, and $W_{\s}U = 0$. Therefore, we may derive a simpler expression for $v$
\begin{align}
    v(\tilde{\z}, \tilde{\s}) = \left(O_1\tilde{\z},  O_2\tilde{\s}\right).
\end{align}
Identifying each of the matrices $O_1$ and $O_2$ as linear maps, we obtain the desired statement. 
\QEDB}

\subsection{Non-identifiability in the Misspecified Non-linear Case}
\label{app:nonlinmisspec}
We begin by stating the counterexample. 
    For $p=p'=1$, $q=1$ and $q'=2$, let us denote $\tilde{z} = z$ and $\tilde{\s} = (s, v)$. We define $f$ as the identity function, and $\tilde{f}$ as the following mixing function:
\begin{equation}
\tilde{f}: \left(\begin{array}{l}
z \\ s \\ v
\end{array}
\right) \mapsto
\left(\begin{array}{c}
z \cdot f(s) +  v \cdot g(s)  \\ s
\end{array}\right),
\end{equation}
with $f(s) = \bm{1}(s \geq 0)$ and $g(s) = \bm{1}(s < 0)$. Below we prove the result for the piecewise constant functions, but following the argument in proof for Example 1, the counter-example holds as long as $g(0) = 0$ and $f(\s) + g(\s) = 1$ almost surely. 

\paragraph{Equality of the target data distributions}
Because $f$ is the identity function, we simply need to show that $\tilde{f}(\tilde{z}, \tilde{\s})$ follows an isotropic Gaussian distribution. We define $u = z \cdot \bm{1}(s \geq 0) +  v \cdot \bm{1}(s < 0)$ and seek to assess the density of $u$ conditionally on $s$. In cases where $s \geq 0$, we get:
\begin{align}
    p(u \mid s, \{s \geq 0\}) &= p_z(u \mid s, \{s \geq 0\})\\
     &= p_z(u).
\end{align}
We proceed similarly for $s < 0$
\begin{align}
    p(u \mid s, \{s < 0\}) &= p_v(u\mid s, \{s < 0\})\\
     &= p_v(u).
\end{align}
Therefore, we have that
\begin{align}
    p(u \mid s) &= \frac{p_v(u)}{2} + \frac{p_z(u)}{2}.
\end{align}
Because $p_v(u)=p_z(u)$ are both the density of the isotropic Gaussian distribution, we conclude that $u$ is independent from $s$ and that $u$ follows an isotropic Gaussian distribution.

\paragraph{Equality of the background data distributions}
Noticing that $\tilde{f}(z, 0, 0) = (z, 0)$, we get that $\tilde{f}(\tilde{z}, \zero) \equald f(z, 0)$.

\paragraph{Entanglement} If we now define $(\bar{z}, \bar{s}) = v(\tilde{z}, \tilde{\s}) = \tilde{f}(\tilde{z}, \tilde{\s})$, we notice that $\bar{z}$ depends non-trivially on $\tilde{\s}$ (via $v$):
\begin{align}
    \bar{z} = z \cdot \bm{1}(s \geq 0) +  v \cdot \bm{1}(s < 0), 
\end{align}
and therefore we have entanglement.

\subsection{Review of Existing Regularization Methods for Comparative Analysis Models}
\label{app:regularization}
We remind the reader of the definition of the aggregated posteriors, defined for each data set,
\begin{align}
    \hat{q}_\phi^t(\z, \s) &= \E_{p_\textrm{data}(\x)}\left[q_\phi(\z \mid \x)q_\phi(\s \mid \x)\right]\\
    \hat{q}_\phi^b(\z, \s) &= \E_{p_\textrm{data}(\x \mid \text{do}(\s =\zero))}\left[q_\phi(\z \mid \x)q_\phi(\s \mid \x)\right].
\end{align}
Additionally, each of those can be used to define marginal distributions over each set of latent variables,
\begin{align}
\hat{q}_\phi^t(\z) &= \E_{p_\textrm{data}(\x)}\left[q_\phi(\z \mid \x)\right]\\
    \hat{q}_\phi^t(\s) &= \E_{p_\textrm{data}(\x)}\left[q_\phi(\s \mid \x)\right]\\
    \hat{q}_\phi^b(\z) &= \E_{p_\textrm{data}(\x \mid \text{do}(\s =\zero))}\left[q_\phi(\z \mid \x)\right].
\end{align}

There are two prominent techniques for promoting disentanglement in the latent space.~\citet{abid2019contrastive} initially proposed to enforce an independence constraint of the form $\hat{q}_\phi^t(\z) \indep \hat{q}_\phi^t(\s)$. Such constraint is enforced using an adversarial classifier that tries to distinguish samples from the joint $\hat{q}^t_\phi(\z, \s)$ from samples from the marginals $\hat{q}^t_\phi(\z), \hat{q}^t_\phi(\s)$ (obtained through random shuffling of the embeddings). 

Later,~\citet{weinberger2022moment} proposed to use a combination of two constraints to help with disentanglement. The first constraint $\hat{q}^t_\phi(\z) = \hat{q}^b_\phi(\z)$ ensures that the distribution of $z$ is identical between the target and the background data set. The second constraint consists in observing that the variational distribution of $\s$ for data points in the background data set is not defined, but in principle could be assessed using the amortization network parameterizing $q_\phi(\s \mid \x)$: 
\begin{align}
    \hat{q}_\phi^b(\s) &= \E_{p_\textrm{data}(\x \mid \text{do}(\s =\zero))}\left[q_\phi(\s \mid \x)\right].
\end{align}
Then, the second constraint is $\hat{q}_\phi^b(\s) = \delta_{\zero}$, where $\delta_{\zero}$ denotes the Dirac distribution centered at zero. In practice, those constraints may be enforced using a maximum mean discrepancy penalty~\citep{gretton12akernel} as used in~\citet{weinberger2022moment}. Another option for enforcing the constraint $\hat{q}_\phi^b(\s) = \delta_{\zero}$ is to penalize with the Wasserstein-2 distance. This second penalization is available in closed-form between a Dirac and a Gaussian distribution~\citep{Villani2008OptimalTO}, as introduced in~\citet{weinberger22disentangling} and used in~\citet{weinberger2023isolating}.

\section{Experiments}

\subsection{Multi-Objective Optimization: the case of Two Objectives}
\label{app:algo}
We recall the formulation of the multi-objective optimization problem
\begin{align}
    \min_{\theta, \phi} \left(-\mathcal{L}^B({\theta, \phi}), -\mathcal{L}^T({\theta, \phi})\right).
\end{align}
The first step in the Multiple-Gradient Descent Algorithm~\citep{desideri2012multiple} consists in calculating the convex combination of the gradients from each loss used for the update of the parameters. In the case of two objective functions, it is defined as:
\begin{align}
        \alpha^*(\theta,\phi) &= \argmin_{\alpha \in [0, 1]} \norm{\alpha\nabla\mathcal{L}^B({\theta, \phi}) + (1-\alpha)\nabla\mathcal{L}^T({\theta, \phi})}_2^2\label{eq:alpha_}.
\end{align}
As pointed out by~\citet{desideri2012multiple}, this optimization problem in Equation~\ref{eq:alpha_} admits a closed-form solution, defined as:
\begin{align}
\label{eq:alpha__}
\alpha^*(\theta, \phi) &= \left[\frac{\left(\nabla\mathcal{L}^T({\theta, \phi}) - \nabla\mathcal{L}^B({\theta, \phi})\right)^\top \nabla\mathcal{L}^T({\theta, \phi})}{\norm{\nabla\mathcal{L}^T({\theta, \phi}) - \nabla\mathcal{L}^B({\theta, \phi})}_2^2}\right]^\tau_{[0, 1]},
\end{align}
where $[a]_{[0, 1]}^\tau = \max(0, \min(a, 1))$ designates the projection onto the compact $[0, 1]$. Taken together, the optimization procedure can be described as:
\begin{align}
    \alpha_t &= \alpha^*(\theta^t, \phi^t) \\
    \delta_t &= -\alpha_t\nabla\mathcal{L}^B({\theta^t, \phi^t}) - (1-\alpha_t)\nabla\mathcal{L}^T({\theta^t, \phi^t})\\
    \left[\theta^{t+1}, \phi^{t+1}\right] &= \left[\theta^{t}, \phi^{t}\right] - \eta \delta_t \label{eq:gradient_up},
\end{align}
where we note that the gradient update in Equation~\ref{eq:gradient_up} may be replaced with that of any first-order stochastic optimizer.

The original framework suggests that one should use the gradient with respect to all the shared parameters of the model in order to compute the alpha parameter in Equation~\ref{eq:alpha__}. This requires collecting the gradients with respect to parameters of every layer of the decoder and the encoder. In the simulations with 
$d_x$ = 150, this amounts to around 180K parameters. For ease of implementation, we calculated $\alpha$ using only the gradients with respect to the weights of the last layer of the decoder (around 20K parameters), and noticed improvements over the single-objective method.

\subsection{Constrained Optimization}
\label{app:optim-constrained}
We aim at solving the following constrained optimization problem:
\begin{align}
    \min_{\theta, \phi} \mathcal{L}({\theta, \phi}) = -\mathcal{L}^B({\theta, \phi}) - \mathcal{L}^T({\theta, \phi}) \text{~~such that~~} \frac{\norm{C_{\z, \s}}_{\text{HS}}^2}{\norm{C_{\z, \z}}_{\text{HS}}\norm{C_{\s, \s}}_{\text{HS}}} \leq \beta.
\end{align}
By using an alternative constraint formulation for the ratio, the problem above is equivalent to: 
\begin{align}
    \min_{\theta, \phi} \mathcal{L}({\theta, \phi}) \text{~~such that~~} \norm{C_{\z, \s}}_{\text{HS}}^2 \leq \beta\norm{C_{\z, \z}}_{\text{HS}}\norm{C_{\s, \s}}_{\text{HS}}.
\end{align}
Finally, utilizing the technique of Lagrange multiplier, we obtain the equivalent problem:
\begin{align}
    \min_{\theta, \phi} \max_{\lambda \geq 0} \mathcal{L}^\lambda({\theta, \phi}) = \mathcal{L}({\theta, \phi}) + \lambda \left(\norm{C_{\z, \s}}_{\text{HS}}^2 - \beta\norm{C_{\z, \z}}_{\text{HS}}\norm{C_{\s, \s}}_{\text{HS}}\right).
\end{align}
To see why this last formulation is equivalent, it is enough to see that when the constraint is not satisfied (i.e., the difference is positive), then the optimal value of the inner optimization problem is $+\infty$ (for $\lambda \rightarrow \infty$), prohibiting those values for the parameters $\theta, \phi$ to be picked by the outer optimization problem. However, the problem is unchanged when the constraint is satified, since the optimal value of the inner optimization problem is $\mathcal{L}({\theta, \phi})$ (for $\lambda \rightarrow 0$. Similar derivations appear in~\citet{gallego2022controlled}.

\subsection{Neural Network Architectures and Implementation Details}
Two separate encoder neural networks were used to parameterize our approximate posterior distributions $q_\phi(\s \mid \x)$ and $q_\phi(\z \mid \x)$. Each network first has a single hidden layer consisting of 128 hidden units, followed by a batch normalization layer, a rectified linear unit (ReLU) activation function, and a dropout layer. Then, the output units were used as inputs to two linear layers, one parameterizing the mean and one parameterizing the log-variance of the variational distribution.  

The decoder network first has two hidden layers with 128 hidden units, taking as input the concatenation of the latent variables $[\z, \s]$, and followed by batch
normalization, a ReLU activation function, and a dropout layer. Then, the outputs units are fed to a linear layer with output size equal to the data dimension, with a softplus activation function.

All models were implemented using PyTorch~\citep{NEURIPS2019_9015} with the scvi-tools framework~\citep{gayoso2022python}. All models were trained for 500 epochs using Adam~\citep{kingma2014adam} with a learning rate of 0.001, using the validation ELBO as an early stopping criterion.


\subsection{Simulation Details}
\label{app:simulation}
We simulate data as follows. We assume we have background measurements (resp. target measurements) from $N_b$ samples (resp. $N_t$ samples). We use $N_t = N_b = 1,500$ throughout the manuscript. 

\paragraph{Target data set} For sample $n$, background latent variable $\z_n$ is generated as:
\begin{align}
 \z_n \sim \textrm{Normal}(0, I_p),
\end{align}
where $p$ is the dimension of the background space. Salient latent variable $\s_n$ is generated as:
\begin{align}
 \s_n \sim \textrm{Normal}(0, I_q),
\end{align}
where $q$ is the dimension of the salient space. Measurements $x_{ng}$ for sample $n$ and feature $g \in \{1, \ldots, G\}$ are generated from a count distribution:
\begin{align}
 x_{ng} \sim \textrm{NegativeBinomial}\left(f^g(\z_n, \s_n), \theta_g \right),
 \label{eq:obs}
\end{align}
where $\theta_g$ is the overdispersion parameter of the negative binomial. We use $G = 150$ throughout the manuscript. When the manuscript mentions Poisson noise, it means that we replace the conditional distribution above by a Poisson distribution, with mean equal to the output of the neural network $f$. The ground truth mixing function $f$ is a neural network with four hidden layers of 40 units, Leaky-ReLU activations with a negative slope of 0.2, and a softmax non-linearity on the last layer to convert the outputs to counts~\citep{lopez2018deep}. The weight matrices of $f$ are sampled according to an isotropic Gaussian distribution, with orthogonal columns, to make sure $f$ is injective~\citep{lachapelle2022disentanglement}. 

\paragraph{Background data set} Proceeding similarly as above, for sample $n$, background latent variable $\z_n$ is generated as:
\begin{align}
 \z_n \sim \textrm{Normal}(0, I_p),
\end{align}
where $p$ is the dimension of the background space. Then, measurements $x_{ng}$ for sample $n$ and feature $g$ are generated from a count distribution:
\begin{align}
 x_{ng} \sim \textrm{NegativeBinomial}\left(f^g(\z_n, \zero), \theta_g \right).
\end{align}

\subsection{Evaluation Metrics}
\label{app:metrics}

\paragraph{Linear Mean Correlation Coefficient (MCC)} 
The Linear Mean Correlation Coefficient (MCC) serves as a metric to evaluate the degree of alignment between inferred and ground-truth latent factors in representation learning, particularly in the context of disentanglement~\citep{khemakhem2020variational}. Specifically, we utilize the mean of the variational posterior $q_\phi(\z, \s \mid \x)$ as an approximation to $\hat{\z}, \hat{\s} = \tilde{f}^{-1}(\x)$. To assess block-wise linear disentanglement, several linear regressions are executed. Initially, to confirm the informativeness of the latent spaces, we:
\begin{itemize}
    \item Predict the ground-truth latent factors $\s$ using the inferred factors $\hat{\s}$ ($\text{MCC}_{\hat{\s}\s})$.
    \item Predict the ground-truth latent factors $\z$ using the inferred factors $\hat{\z}$ ($\text{MCC}_{\hat{\z}\z})$.
\end{itemize}
Subsequently, to ensure the absence of undesired overlaps, we:
\begin{itemize}
    \item Predict the ground-truth latent factors $\s$ using the inferred factors $\hat{\z}$ ($\text{MCC}_{\hat{\z}\s})$.
    \item Predict the ground-truth latent factors $\z$ using the inferred factors $\hat{\s}$ ($\text{MCC}_{\hat{\s}\z})$.
\end{itemize}
For each set of predicted latent factors, the Pearson (and equivalently, Spearman) linear MCC is computed as the mean Pearson (or Spearman) Correlation Coefficient between predictions and the ground truth, evaluated component-wise. The MCC value ranges between -1 and 1, with values closer to 1 indicating a strong positive linear relationship, values closer to -1 indicating a strong negative linear relationship, and values around 0 indicating no linear relationship.

\paragraph{Average Silhouette Width (AWS)}
This metric assumes at disposal an Euclidean space where each data point $n$ is associated with an embedding vector $t_n \in \R^d$ where $n$ is a data point. Additionally, the AWS requires the definition of cluster assignments $y_n \in \{1, \ldots, K\}$, where $K$ is the total number of clusters. For each data point $n$, the we define silhouette score $\text{SS}_n$ of sample $n$ as 
\begin{align}
    \text{SS}_n = \frac{b_n-a_n}{\max{(a_n, b_n)}},
\end{align}
where $a_n$ is the average distance between data point $n$ and all of other points with the same cluster label, and $b_n$ is the average distance between $n$ and all the points in the next nearest cluster. Then, for a data set with $N$ samples, the AWS is defined as:
\begin{align}
    \text{ASW} = \frac{1}{N}\sum_{n=1}^N \text{SS}_n.
\end{align}
The value of ASW lies between -1 and 1, where a higher value indicates a better ability to distinguish the clusters in the embedding space.

\paragraph{Adjusted Rand Index (ARI)}
The Adjusted Rand Index (ARI) is a metric used to measure the similarity between two data clusterings. Consider two sets of cluster assignments: the true assignments \(y_n \in \{1, \ldots, K\}\) and the predicted assignments \(y'_n \in \{1, \ldots, K'\}\), where \(n\) is a data point, \(K\) is the total number of true clusters, and \(K'\) is the total number of predicted clusters. The ARI takes into account the number of pairings of data points that are in the same or different clusters for both the true and predicted assignments. Specifically, the ARI is defined as:
\begin{align}
    \text{ARI} = \frac{\sum_{ij} \binom{n_{ij}}{2} - \left[ \sum_{i} \binom{a_i}{2} \sum_{j} \binom{b_j}{2} \right] / \binom{N}{2}}{\frac{1}{2} \left[ \sum_{i} \binom{a_i}{2} + \sum_{j} \binom{b_j}{2} \right] - \left[ \sum_{i} \binom{a_i}{2} \sum_{j} \binom{b_j}{2} \right] / \binom{N}{2}},
\end{align}
where \( n_{ij} \) is the number of data points that are both in cluster \( i \) of the true assignments and cluster \( j \) of the predicted assignments. \( a_i \) and \( b_j \) are the total number of data points in cluster \( i \) of the true assignments and cluster \( j \) of the predicted assignments, respectively. \( N \) is the total number of data points. The value of ARI lies between -1 and 1, where a higher value indicates better clustering performance.

\paragraph{Normalized Mutual Information (NMI)}
The Normalized Mutual Information (NMI) is a metric designed to gauge the similarity between two data clusterings. Given two sets of cluster assignments: the true assignments \(y_n \in \{1, \ldots, K\}\) and the predicted assignments \(y'_n \in \{1, \ldots, K'\}\), where \(n\) is a data point, \(K\) is the total number of true clusters, and \(K'\) is the number of predicted clusters, the NMI is computed using the mutual information (MI) between the two assignments and their respective entropies:

\begin{align}
    \text{NMI}(y, y') = \frac{2 \times \text{MI}(y, y')}{H(y) + H(y')}.
\end{align}

Mutual information between the true and predicted assignments is given by:
\begin{align}
    \text{MI}(y, y') = \sum_{i=1}^{K} \sum_{j=1}^{K'} p(i, j) \log \left( \frac{p(i, j)}{p(i)p'(j)} \right),
\end{align}
where \( p(i, j) \) is the joint probability of a data point belonging to cluster \(i\) in the true assignments and cluster \(j\) in the predicted assignments. \( p(i) \) and \( p'(j) \) are the probabilities of a data point belonging to cluster \(i\) and \(j\) in the true and predicted assignments, respectively.

The entropies of the true and predicted assignments are:
\begin{align}
    H(y) &= -\sum_{i=1}^{K} p(i) \log(p(i)), \\
    H(y') &= -\sum_{j=1}^{K'} p'(j) \log(p'(j)).
\end{align}

The value of NMI lies in the range [0, 1], with a score of 1 suggesting that the two sets of cluster assignments are identical, while a score of 0 indicates no shared information between them.

\subsection{Baseline Models}
\label{app:baselines}
All baselines share the same architecture and implementations for the sake of comparison. All are modifications of the code from the ContrastiveVI package~\citep{weinberger2023isolating}. Below we provide additional details about how each baseline was utilized.

\paragraph{VAE} Because a VAE does not have two latent spaces, in all experiments, we double the number of latent variables to match the one from our contrastive analysis models. Then, we assign each latent variable to either the background or the salient space. Our assignment method consists in applying cPCA~\citep{abid2018exploring} to the learned representations from the VAE to split the latent space into a background and a salient space. As a first attempt, we used the loadings from the top cPCA eigenvalues to project the latent space into a salient space, and then used the loadings from the top PCA eigenvalue of the background data set to project the latent space into a background space. This approach had extremely poor performance ($\delta$-MCC close to zero in most experiments, because of high cross MCC). Therefore, we proceeded as follows: We use the loadings onto the first contrastive principal component to obtain the list of latent units that contribute most to explaining the differences between the target and the background data set (using the absolute value of the eigenvector). We assign variables with the highest score to the salient space, and the ones with the lowest score to the background space. 


\paragraph{ContrastiveVI} Our ContrastiveVI implementation consists of exactly the SO-U-cVAE method with the addition of the Wasserstein penalty to the ELBO. More precisely, if $\mu_{\s}(\x)$ and $\sigma_{\s}(\x)$ encode the mean and standard deviation parameter of $q_\phi(\s \mid \x)$, respectively, then the Wasserstein penalty $\mathcal{L}_\mathcal{W}$ is derived as
\begin{align}
    \mathcal{L}_\mathcal{W} = \norm{\mu_{\s}(\x)}_2^2 + \norm{\sigma_{\s}(\x)}^2_2,
\end{align}
with a fixed hyperparameter for the regularization strength (i.e., multiplier equal to one).

\paragraph{CausalDiscrepancyVAE}  CausalDiscrepancyVAE~\citep{zhang2023identifiability} is generative model with a latent causal graph (DAG), a polynomial mixing function, and an interventional model for single and double-node (soft) interventions. Because~\citet{zhang2023identifiability} also apply their method to the data set from~\citet{norman2019exploring}, we discuss this work here. We note that those two models have distinct purposes. Contrastive Analysis methods aim at learning informative representation of the perturbations by removing signal from the heterogeneity of the control population. However, CausalDiscrepancyVAE aims at learning a DAG in latent space. For that, CausalDiscrepancyVAE requires the observation of many interventional regimes, as well as knowledge of the targets per intervention (ignored during our benchmark). For this reason, the models are not entirely comparable. Although CausalDiscrepancyVAE has a rigorous causal semantic, but it does not consider modeling of background latent variables, which from our experience, this is however necessary to get high-quality embeddings of the interventional data. To illustrate this point, we assessed how well the embedding from CausalDiscrepancyVAE may reflect biological information, using our benchmark (ARI, NMI, ASW of the different pathways captured by the experiment). Using the public code and available model trained by the authors (105 latent variables). We embedded all cells using the encoder network (after the DAG layer) and found that the performance was poor. Because our evaluation may be dependent on the number of latent variables, we re-tried this with a model that we fit ourselves from the available code, this time with 20 latent variables, and obtained similar results.

\subsection{Real-word Data Details}

\paragraph{Data Preprocessing} We downloaded the data set using the ContrastiveVI package~\citep{weinberger2023isolating}. The data set has measurements of the effects on gene expression levels of 284 different CRISPR-mediated perturbations on K562 cells. Each perturbation induced overexpression of a single gene or a pair of genes. The background data set ($8,907$ cells) is defined as all unperturbed cells. The target data set ($24,913$ cells) is defined as all the perturbed cells whose genetic perturbation was labelled with a pathway by the authors of the original manuscript~\citep{norman2019exploring}. The labeled pathways in the dataset are G1 Cycle, Erythroid, Pioneer Factors, Granulocyte Apoptosis, Megakaryocyte, and Pro Growth. We also filtered genes to retain the top $2,000$ highly variable genes. 

\paragraph{Number of latent variables}
For our main results, we used 10 dimensional $\z$ and $\s$ .

\subsection{Qualitative Comparison of Methods on Real-World Data}
\label{app:add-results}

\begin{figure}[ht]
    \centering
    \includegraphics[width=\textwidth]{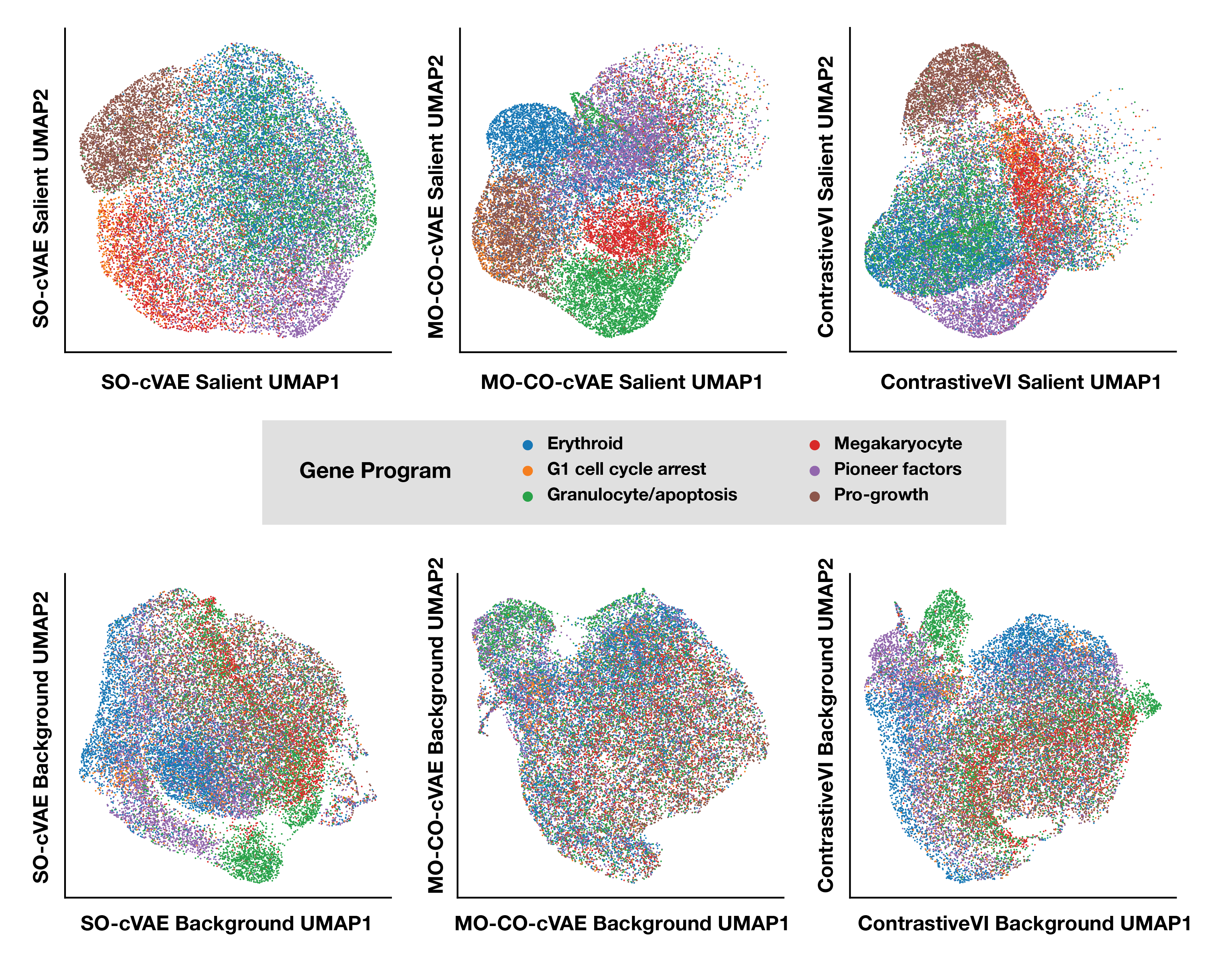}
    \caption{UMAP visualization of salient and background spaces from SO-cVAE, MO-CO-cVAE, as well as ContrastiveVI.  Each point is a
cell. Cells are colored by their group of genetic perturbation, where groups were assigned based on biological annotation from the authors of ~\citet{norman2019exploring}.}
    \label{fig:UMAPs}
\end{figure}

To understand the impact of the differences in metrics we reported in Table~\ref{tab:real}, we visualized the learned latent spaces for SO-cVAE, MO-CO-cVAE, as well as ContrastiveVI with UMAP (Figure~\ref{fig:UMAPs}). 

\paragraph{Salient space interpretation} In accordance with our desiderata, we observe for each method that the salient space recapitulates the perturbation type more accurately than the background space. However, we clearly notice that MO-CO-cVAE has best performance, as it more clearly delineates cells that underwent perturbations from the group ``erythoid'' versus ``granulocytes/apoptosis''. This was suggested by the high clustering and silhouette scores in Table~\ref{tab:real}.

\paragraph{Background space interpretation} Additionally, we would like to make sure that the latent space contains as few information about the perturbations as possible. In this regard, it is important to note that, as expected, the un-regularized model SO-cVAE shows important separation of perturbation groups in its background latent space (as suggested by the high cross MCC in Table~\ref{tab:real}). However, the other methods show lesser leakage of that information in their background space. 

Importantly, we see that the apoptosis group is still separated even in our model, which performs best in this benchmark. We attribute this to the fact that perturbations may sometimes bias cells towards expressing some gene programs that are inherently varying in the control population (such as cell death, here). Therefore, it is impossible to expect that $\z$ will be perfectly independent from the perturbation label on this data set.

\subsection{Additional Results on Simulated Data}
\label{app:results_supp}

\paragraph{Comparison to MultiDomainCRL} In order to investigate the performance of the ICA source matching method from~\cite{sturma2023unpaired}, we applied their publicly available code on our simulated data, where the background data set and the target data set are used as two domains. We applied the FastICA algorithm with the known number of sources in each domain (e.g., 5 for the background data set, and 5 + 5 = 10 for the target data set). Then, we applied their individual source matching procedure, expecting it to match all the sources from the background data to one of the sources in the target data. Interestingly, the method consistently miss-identified the number of shared sources (underestimation), failing to match some of the background sources. This is already suggestive that the method may not work well on our benchmark. To report performance in a systematic fashion, we slightly modified their matching algorithm to associate each source in the background to exactly one source of the target data. The remainder of the sources for the target data set constituted the inferred salient variables, whereas the inferred salient variables for the background were set to zero. Even though we tried different metrics for matching latent variables (Wasserstein distance, as well as Smirnov Two-Sample Test) and also different strategies for normalizing the input data (raw values and logarithmic scale), the $\delta$-MCC in the simple scenario of Table~\ref{tab:simu_mcc} was systematically between -0.1 and 0.1 for all experiments.

\paragraph{Misspecification of activation function} We investigated how the performance changed when the data generating process was altered so that the leaky ReLU activation function is replaced by a hyperbolic tangent activation function (but the model stays the same). More precisely, we generated data according to two regimes. In the first one, the scale of the values at the hidden layer is small ($\ll$ 1). This is interesting because in this case the tanh function well approximated a linear function (referred to as quasi-linear). In the second case, the scale of the values is larger ($\sim$1), and the tanh function is effectively non-linear (referred to as non-linear). For both data sets, we ran the method SO-cVAE in the ideal setting to assess disentanglement (akin to Table~\ref{tab:simu_mcc}). We report those results in Table~\ref{tab:mispec_activation}. As expected, the performance is high in the quasi-linear case, and indeed it surpasses the results of the manuscript. This is in agreement with the intuition that a linear model is easier to identify from data compared to a piecewise-linear model. However, the performance drops in the non-linear case, which illustrates that identifiability is harder in this case. (It could also be explained by the mismatch between the data and the model, or by the excessive saturation from the tanh function at input values $\gg$ 1). 
\input{tables/mispec_activation}

\paragraph{Additional simulation results} In this section, we present the following experimental results:
\begin{itemize}
    \item The Pearson MCC scores under Gaussian noise when the number of latent variable is known (Table~\ref{tab:simu_mcc_gaussian})
    \item The Spearman MCC scores under Poisson and negative binomial noise when the number of latent variable is known (Table~\ref{tab:simu_mcc_appendix}). 
    \item The Pearson MCC scores when the number of latent variables is unknown (Table~\ref{tab:mispec_app}).
    \item The Spearman MCC scores when the number of latent variables is unknown (Table~\ref{tab:mispec_app_spearman}).
    \item The Pearson MCC scores under regularization (Table~\ref{tab:mispec_regu_app}).
    \item The Spearman MCC scores under regularization (Table~\ref{tab:mispec_regu_app_spearman}).
\end{itemize}

\input{tables/simu_mcc_gaussian}

\input{tables/simu_mcc_appendix}

\input{tables/simu_misspecification_supp}

\input{tables/simu_regularization_appendix}

%% file: 7_relatedwork.tex
\section{Related Work}
\label{app:related-work}

\paragraph{Interventional Causal Representation Learning} Several recent works have investigated the setting of learning from multiple data sets with interventional shifts in latent space. For example, \citet{lachapelle2022disentanglement} proposed an identifiable non-linear ICA based on the assumption that a rich set of interventional data is available, where each intervention shifts the sparse set of sufficient statistics of the latent variable prior distribution. \citet{lopez2023learning} proposed an application of this theory to the setting of modeling single-cell perturbation data, and reported empirical evidence, based on simulations, that the identifiability guarantee might hold for count data. Our work is distinct from these as it considers the setting of a small number of data sets, and is mainly concerned with subspace identification, but it does provide a first line of attack towards extending the results from~\citet{lachapelle2022disentanglement} for counting observational noise. \citet{ahuja2023interventional} recently proposed a framework for proving identifiability of noiseless auto-encoders under the assumption of a large set of interventions, and a polynomial decoder. \citet{buchholz2023learning,von2023nonparametric,jiang2023learning} are concerned with the identifiability of non-linear ICA, under interventional data and with a general class of non-linear mixing functions (either parametric, or non-parametric). However, the assumptions made by these works are restricted to stochastic interventions, which makes them not applicable to our problem. Interestingly, Lemma 8, Appendix D.1 from~\citet{buchholz2023learning} points out that non-stochastic interventions create some form of unidentifiability. Consequently, they did not study it in detail (unlike our work). Also, such works require the availability of data from as many interventions as the number of the latent dimensions, while our work solely considers two separate environments. 

\paragraph{Identifiability of Modular Representations} Several recent works specifically investigated the prospect of block-wise identifiability. \citet{lachapelle2022partial} proved that under a relaxation of the assumptions from~\cite{lachapelle2022disentanglement}, we may obtain only disentanglement by block (when interventions do not dissect enough the latent space to recover the ground truth mixing function). \citet{lachapelle2023additive} investigated the setting of additive decoders, where each decoder uses only a block of latent variables. In this setting, the goal is to prove the block-wise identifiability of the latent variables. The definitions of block-wise identifiability in these recent works~\citep{lachapelle2023additive,von2021self} are essentially equivalent to the ones considered in this manuscript. \citet{kong22apartial} applied non-linear ICA theory to the domain adaptation problem, and showed block-wise identifiability of the effect of the domain with the predicted outcome under their latent variable model.

\paragraph{Source Matching Across Domains}
In the classical linear ICA problem, we are interested in learning $z$ from data generated as $x = Wz + \epsilon$. Framing the contrastive analysis problem in the paradigm of linear ICA, we would observe background data $x^b = W^bz +\epsilon^b$ as well as target data $x^t = W^ts + W^bz + \epsilon^t$. The target data set has been generated with additional sources $s$ that we would like to identify collectively and to separate from the background sources $z$. In the case of genomics, the parameters $W^t$ are also relevant, as they encode which genes are associated with which components of the novel sources $s$. For example, in~\citet{boileau2020exploring} the sparse entries of the matrix 
$W^t$ are used to identify genes associated with leukemia. The problem of matching sources across different ICA models is treated in~\citet{sturma2023unpaired}, although in the context of the more general problem, in which the two data sets may be composed of different observable quantities (= modalities). Their solution considers an idealized scenario with a linear mixing function, and no observation noise, but could provide a reasonable baseline derived from linear ICA. We found that the performance of the method was not competitive on the simulated data, likely because the mixing function used for generating the data is not linear. 

%% file: tables/mispec_activation.tex
\begin{table}[ht]
\centering
\caption{Identifiability under assumptions of known dimensions of latent spaces. Pearson MCC for SO-cVAE.}\vspace{2mm}
\begin{tabular}{cccccc}
\toprule
    \textbf{Data set}                                                                                  & \textbf{MCC}$_{\hat{\z}\z}$ ($\uparrow$) & \textbf{MCC}$_{\hat{\z}\s}$ ($\downarrow$) & \textbf{MCC}$_{\hat{\s}\z}$ ($\downarrow$) & \textbf{MCC}$_{\hat{\s}\s}$ ($\uparrow$) & $\delta$\textbf{-MCC} ($\uparrow$) \\
\midrule
 {\footnotesize \begin{tabular}[x]{@{}c@{}}\textbf{Tanh}\\\textbf{Quasi-linear}\end{tabular} }& $0.956 \pm 0.001$ & $0.047 \pm 0.008$ & $0.051 \pm 0.005$ & $0.963 \pm 0.002$ & $0.910 \pm 0.007$ \\ \midrule
 {\footnotesize \begin{tabular}[x]{@{}c@{}}\textbf{Rescaled Tanh}\\\textbf{Non-linear}\end{tabular}} & $0.834 \pm 0.011$ & $0.122 \pm 0.020$ & $0.139 \pm 0.033$ & $0.541 \pm 0.041$ & $0.557 \pm 0.043$ \\
\bottomrule
\label{tab:mispec_activation}
\end{tabular}
\end{table}

%% file: tables/simu_mcc_gaussian.tex
\begin{table}[p]
\centering
\caption{Identifiability under assumptions of known dimensions of latent spaces. Best in bold.}\vspace{2mm}
\begin{tabular}{lc|ccccc}
\toprule
     \textbf{Model}         &     \textbf{Noise}                                                                                  & \textbf{MCC}$_{\hat{\z}\z}$ ($\uparrow$) & \textbf{MCC}$_{\hat{\z}\s}$ ($\downarrow$) & \textbf{MCC}$_{\hat{\s}\z}$ ($\downarrow$) & \textbf{MCC}$_{\hat{\s}\s}$ ($\uparrow$) & $\delta$\textbf{-MCC} ($\uparrow$) \\\midrule
\textbf{SO-cVAE} & \multirow{3}{*}{\textbf{Gaussian}} & $\mathbf{0.96 \pm 0.01}$ & $\mathbf{0.12 \pm 0.03}$ & $\mathbf{0.08 \pm 0.01}$ & $\mathbf{0.96 \pm 0.01}$ & $\mathbf{0.86 \pm 0.01}$ \\
\textbf{MO-cVAE} & & $\mathbf{0.96 \pm 0.01}$ & $0.13 \pm 0.03$ & $\mathbf{0.08 \pm 0.02}$ & $\mathbf{0.96 \pm 0.01}$ & $\mathbf{0.86 \pm 0.01}$ \\
\textbf{VAE}     & & $\mathbf{0.96 \pm 0.01}$ & $0.15 \pm 0.02$ & $0.15 \pm 0.01$ & $0.95 \pm 0.01$ & $0.80 \pm 0.02$ \\
\bottomrule
\label{tab:simu_mcc_gaussian}
\end{tabular}
\end{table}

%% file: tables/simu_mcc_appendix.tex
\begin{table}[ht]
\centering
\caption{Identifiability of contrastive analysis models under assumptions of known dimensions of latent spaces (Spearman MCC).\label{tab:simu_mcc_appendix}}
\scalebox{0.9}{
\begin{tabular}{cc|ccccc}
\toprule
     \textbf{Model}         &     \textbf{Noise}   &                                                                             \textbf{MCC}$_{\hat{\z}\z}$ ($\uparrow$) & \textbf{MCC}$_{\hat{\z}\s}$ ($\downarrow$) & \textbf{MCC}$_{\hat{\s}\z}$ ($\downarrow$) & \textbf{MCC}$_{\hat{\s}\s}$ ($\uparrow$) & $\delta$\textbf{-MCC} ($\uparrow$)  \\\midrule
\textbf{MO-cVAE} & \multirow{3}{*}{\textbf{Poisson}}                                                     &   $\mathbf{0.92 \pm 0.01}$           &   $\mathbf{0.08 \pm 0.01}$            &    $\mathbf{0.07 \pm 0.02}$            &    $\mathbf{0.96 \pm 0.01}$            &          $\mathbf{0.87 \pm 0.01}$     \\
\textbf{SO-cVAE} &            &   $\mathbf{0.92 \pm 0.01}$           &   $\mathbf{0.08 \pm 0.01}$            &    $0.08 \pm 0.02$            &    $\mathbf{0.96 \pm 0.01}$            &          $0.86 \pm 0.01$     \\
\textbf{VAE}  &  & $0.88 \pm 0.04$           &   $0.17 \pm 0.09$            &    $0.14 \pm 0.07$            &    $0.94 \pm 0.04$            &          $0.76 \pm 0.12$     \\ \midrule
\textbf{MO-cVAE} & \multirow{3}{*}{\textbf{\begin{tabular}[c]{@{}l@{}}Negative\\ binomial\end{tabular}}} &      $\mathbf{0.95 \pm 0.01}$           &   $0.14 \pm 0.01$            &    $\mathbf{0.07 \pm 0.01}$            &    $\mathbf{0.96 \pm 0.01}$            &          $\mathbf{0.84 \pm 0.01}$     \\
\textbf{SO-cVAE} &  &      $\mathbf{0.95 \pm 0.01}$           &   $\mathbf{0.13 \pm 0.02}$            &    $\mathbf{0.07 \pm 0.01}$            &    $0.95 \pm 0.01$            &          $\mathbf{0.84 \pm 0.01}$     \\
\textbf{VAE}   & &      $0.82 \pm 0.11$           &   $0.46 \pm 0.18$            &    $0.37 \pm 0.18$            &    $0.82 \pm 0.10$            &          $0.41 \pm 0.28$     \\
\bottomrule
\end{tabular}
}
\end{table}

%% file: tables/simu_misspecification_supp.tex
\begin{table}[ht]
\centering
\caption{Identifiability of contrastive analysis models under misspecification of latent dimensions (Pearson MCC).\label{tab:mispec_app}}
\scalebox{0.9}{
\begin{tabular}{lllllll}
\toprule
  & $q$        & \textbf{MCC}$_{\hat{\z}\z}$ ($\uparrow$) & \textbf{MCC}$_{\hat{\z}\s}$ ($\downarrow$) & \textbf{MCC}$_{\hat{\s}\z}$ ($\downarrow$) & \textbf{MCC}$_{\hat{\s}\s}$ ($\uparrow$) & $\delta$\textbf{-MCC} ($\uparrow$)  \\\midrule
\multirow{4}{*}{\textbf{SO-cVAE}}&\textbf{5} &            $\mathbf{0.91 \pm 0.01}$           &   $\mathbf{0.08 \pm 0.01}$            &    $\mathbf{0.07 \pm 0.02}$            &    $0.92 \pm 0.01$            &          $0.84 \pm 0.01$  \\
&\textbf{7}  & $\mathbf{0.91 \pm 0.01}$ & $\mathbf{0.08 \pm 0.01}$ & $0.30 \pm 0.06$ & $\mathbf{0.94 \pm 0.02}$ & $0.73 \pm 0.03$ \\
&\textbf{10} & $\mathbf{0.91 \pm 0.01}$ & $\mathbf{0.08 \pm 0.01}$ & $0.45 \pm 0.02$ & $\mathbf{0.94 \pm 0.02}$ & $0.66 \pm 0.01$ \\
&\textbf{15} & $\mathbf{0.91 \pm 0.01}$ & $\mathbf{0.08 \pm 0.02}$ & $0.59 \pm 0.04$ & $\mathbf{0.94 \pm 0.02}$ & $0.58 \pm 0.02$ \\ \midrule
\multirow{4}{*}{\textbf{MO-cVAE}}&\textbf{5}  &   $\mathbf{0.91 \pm 0.01}$           &   $\mathbf{0.08 \pm 0.01}$            &    $\mathbf{0.07 \pm 0.01}$            &    $\mathbf{0.94 \pm 0.01}$            &          $\mathbf{0.85 \pm 0.01}$ \\
&\textbf{7}  & $\mathbf{0.91 \pm 0.01}$ & $\mathbf{0.08 \pm 0.01}$ & $0.15 \pm 0.02$ & $\mathbf{0.94 \pm 0.02}$ & $0.81 \pm 0.01$ \\
&\textbf{10} & $\mathbf{0.91 \pm 0.01}$ & $0.09 \pm 0.01$ & $0.25 \pm 0.03$ & $\mathbf{0.94 \pm 0.02}$ & $0.75 \pm 0.01$ \\
&\textbf{15} & $\mathbf{0.91 \pm 0.01}$ & $\mathbf{0.08 \pm 0.02}$ & $0.36 \pm 0.04$ & $\mathbf{0.94 \pm 0.02}$ & $0.70 \pm 0.02$\\
\bottomrule
\end{tabular}
}
\end{table}


\begin{table}[ht]
\centering
\caption{Identifiability of contrastive analysis models under misspecification of latent dimensions (Spearman MCC).\label{tab:mispec_app_spearman}}
\scalebox{0.9}{
\begin{tabular}{lllllll}
\toprule
     & $q$      & \textbf{MCC}$_{\hat{\z}\z}$ ($\uparrow$) & \textbf{MCC}$_{\hat{\z}\s}$ ($\downarrow$) & \textbf{MCC}$_{\hat{\s}\z}$ ($\downarrow$) & \textbf{MCC}$_{\hat{\s}\s}$ ($\uparrow$) & $\delta$\textbf{-MCC} ($\uparrow$)  \\ \midrule
\multirow{4}{*}{\textbf{SO-cVAE}} & \textbf{5}  & 0.92 $\pm$ 0.01 & 0.08 $\pm$ 0.01 & \textbf{0.08 $\pm$ 0.02} & \textbf{0.96 $\pm$ 0.01} & 0.86 $\pm$ 0.01 \\
&\textbf{7}  & \textbf{0.93 $\pm$ 0.01} & \textbf{0.08 $\pm$ 0.01} & 0.30 $\pm$ 0.06 & \textbf{0.96 $\pm$ 0.00} & 0.75 $\pm$ 0.03 \\
&\textbf{10} & \textbf{0.93 $\pm$ 0.01} & \textbf{0.08 $\pm$ 0.01} & 0.45 $\pm$ 0.03 & \textbf{0.96 $\pm$ 0.00} & 0.68 $\pm$ 0.02 \\
&\textbf{15} & \textbf{0.93 $\pm$ 0.01} & \textbf{0.08 $\pm$ 0.02} & 0.60 $\pm$ 0.03 & \textbf{0.96 $\pm$ 0.00} & 0.60 $\pm$ 0.02 \\ \midrule
\multirow{4}{*}{\textbf{MO-cVAE}}&\textbf{5}   & 0.92 $\pm$ 0.01 & \textbf{0.08 $\pm$ 0.01} & \textbf{0.07 $\pm$ 0.02} & \textbf{0.96 $\pm$ 0.01} & \textbf{0.87 $\pm$ 0.01} \\
&\textbf{7}  & \textbf{0.93 $\pm$ 0.01} & \textbf{0.08 $\pm$ 0.01} & 0.15 $\pm$ 0.02 & \textbf{0.96 $\pm$ 0.00} & 0.83 $\pm$ 0.01 \\
&\textbf{10} & \textbf{0.93 $\pm$ 0.01} & \textbf{0.08 $\pm$ 0.01} & 0.26 $\pm$ 0.04 & \textbf{0.96 $\pm$ 0.00} & 0.77 $\pm$ 0.02 \\
&\textbf{15} & \textbf{0.93 $\pm$ 0.01} & \textbf{0.08 $\pm$ 0.02} & 0.37 $\pm$ 0.04 & \textbf{0.96 $\pm$ 0.00} & 0.72 $\pm$ 0.02 \\
\bottomrule
\end{tabular}
}
\end{table}

%% file: tables/simu_regularization_appendix.tex
\begin{table}[ht]
\centering
\caption{The impact of regularization on contrastive analysis models under misspecification of latent dimensions (Pearson MCC).\label{tab:mispec_regu_app}}
\scalebox{0.9}{
\begin{tabular}{lllllll}\toprule
       &     & \textbf{MCC}$_{\hat{\z}\z}$ ($\uparrow$) & \textbf{MCC}$_{\hat{\z}\s}$ ($\downarrow$) & \textbf{MCC}$_{\hat{\s}\z}$ ($\downarrow$) & \textbf{MCC}$_{\hat{\s}\s}$ ($\uparrow$) & $\delta$\textbf{-MCC} ($\uparrow$)  \\
\midrule
\multirow{4}{*}{\textbf{SO-U-cVAE}} &$\lambda=0$ & $\mathbf{0.91 \pm 0.01}$ & $0.08 \pm 0.01$ & $0.45 \pm 0.02$ & $\mathbf{0.94 \pm 0.00}$ & $0.66 \pm 0.01$ \\
&$\lambda = 10$ & $\mathbf{0.91 \pm 0.01}$ & $0.08 \pm 0.01$ & $0.37 \pm 0.04$ & $\mathbf{0.94 \pm 0.00}$ & $0.70 \pm 0.02$ \\
&$\lambda = 50$ & $\mathbf{0.91 \pm 0.01}$ & $0.08 \pm 0.01$ & $0.25 \pm 0.03$ & $\mathbf{0.94 \pm 0.00}$ & $0.76 \pm 0.01$ \\
&$\lambda = 100$ & $0.80 \pm 0.03$ & $0.11 \pm 0.05$ & $0.29 \pm 0.08$ & $0.92 \pm 0.03$ & $0.66 \pm 0.08$ \\ \midrule
\textbf{SO-CO-cVAE}& & $\mathbf{0.91 \pm 0.01}$ & $\mathbf{0.07 \pm 0.01}$ & $0.23 \pm 0.03$ & $\mathbf{0.94 \pm 0.00}$ & $0.77 \pm 0.01$ \\ \midrule
\multirow{4}{*}{\textbf{MO-U-cVAE}}  & $\lambda = 0$ & $\mathbf{0.91 \pm 0.01}$ & $0.09 \pm 0.01$ & $0.25 \pm 0.03$ & $\mathbf{0.94 \pm 0.00}$ & $0.75 \pm 0.01$ \\
&$\lambda = 10$ & $\mathbf{0.91 \pm 0.01}$ & $0.08 \pm 0.01$ & $0.21 \pm 0.02$ & $\mathbf{0.94 \pm 0.00}$ & $0.78 \pm 0.01$ \\
&$\lambda = 50$ & $0.88 \pm 0.03$ & $0.08 \pm 0.01$ & $0.16 \pm 0.02$ & $\mathbf{0.94 \pm 0.00}$ & $0.79 \pm 0.02$ \\
&$\lambda = 100$ & $0.79 \pm 0.02$ & $0.09 \pm 0.01$ & $0.19 \pm 0.02$ & $0.93 \pm 0.00$ & $0.73 \pm 0.03$ \\ \midrule
\textbf{MO-CO-cVAE}& & $\mathbf{0.91 \pm 0.01}$ & $0.08 \pm 0.01$ & $\mathbf{0.17 \pm 0.02}$ & $\mathbf{0.94 \pm 0.00}$ & $\mathbf{0.80 \pm 0.01}$ \\ \bottomrule
\end{tabular}
}
\end{table}

\begin{table}[ht]
\centering
\caption{The impact of regularization on contrastive analysis models under misspecification of latent dimensions (Spearman MCC).\label{tab:mispec_regu_app_spearman}}
\scalebox{0.9}{
\begin{tabular}{lllllll}\toprule
       &     & \textbf{MCC}$_{\hat{\z}\z}$ ($\uparrow$) & \textbf{MCC}$_{\hat{\z}\s}$ ($\downarrow$) & \textbf{MCC}$_{\hat{\s}\z}$ ($\downarrow$) & \textbf{MCC}$_{\hat{\s}\s}$ ($\uparrow$) & $\delta$\textbf{-MCC} ($\uparrow$)  \\
\midrule
\multirow{4}{*}{\textbf{SO-U-cVAE}}  &$\lambda = 0$ & $\mathbf{0.93 \pm 0.01}$ & $0.08 \pm 0.01$ & $0.45 \pm 0.03$ & $\mathbf{0.96 \pm 0.00}$ & $0.68 \pm 0.02$ \\
&$\lambda = 10$ & $\mathbf{0.93 \pm 0.01}$ & $0.08 \pm 0.01$ & $0.38 \pm 0.04$ & $\mathbf{0.96 \pm 0.00}$ & $0.71 \pm 0.02$ \\
&$\lambda = 50$ & $0.92 \pm 0.01$ & $0.08 \pm 0.01$ & $0.25 \pm 0.02$ & $\mathbf{0.96 \pm 0.00}$ & $0.78 \pm 0.01$ \\
&$\lambda = 100$ & $0.81 \pm 0.03$ & $0.11 \pm 0.06$ & $0.29 \pm 0.08$ & $0.94 \pm 0.03$ & $0.67 \pm 0.08$ \\ \midrule
\textbf{SO-CO-cVAE} & & $\mathbf{0.93 \pm 0.01}$ & $\mathbf{0.07 \pm 0.01}$ & $0.23 \pm 0.03$ & $\mathbf{0.96 \pm 0.00}$ & $0.79 \pm 0.01$\\ \midrule
\multirow{4}{*}{\textbf{MO-U-cVAE}}&$\lambda = 0$ & $\mathbf{0.93 \pm 0.01}$ & $0.08 \pm 0.01$ & $0.26 \pm 0.04$ & $\mathbf{0.96 \pm 0.00}$ & $0.77 \pm 0.02$ \\
&$\lambda = 10$ & $\mathbf{0.93 \pm 0.01}$ & $0.08 \pm 0.01$ & $0.22 \pm 0.03$ & $\mathbf{0.96 \pm 0.00}$ & $0.79 \pm 0.01$ \\
&$\lambda = 50$ & $0.89 \pm 0.03$ & $0.08 \pm 0.01$ & $0.16 \pm 0.03$ & $\mathbf{0.96 \pm 0.00}$ & $0.80 \pm 0.03$ \\
&$\lambda = 100$ & $0.81 \pm 0.02$ & $0.08 \pm 0.01$ & $0.20 \pm 0.02$ & $\mathbf{0.96 \pm 0.00}$ & $0.74 \pm 0.03$ \\ \midrule
\textbf{MO-CO-cVAE} & & $\mathbf{0.93 \pm 0.01}$ & $0.08 \pm 0.01$ & $\mathbf{0.18 \pm 0.02}$ & $\mathbf{0.96 \pm 0.00}$ & $\mathbf{0.82 \pm 0.01}$ \\ \bottomrule
\end{tabular}
}
\end{table}